\documentclass[a4paper,fleqn]{cas-dc}

\usepackage[authoryear,longnamesfirst]{natbib}
\usepackage{graphicx}%
\usepackage{subcaption}%
\usepackage{multirow}%
\usepackage{amsmath,amssymb,amsfonts}%
\usepackage{amsthm}%
\usepackage{mathrsfs}%
\usepackage[title]{appendix}%
\usepackage{xcolor}%
\usepackage{textcomp}%
\usepackage{manyfoot}%
\usepackage{booktabs}%
\usepackage{caption}%
\usepackage{algorithm}%
\usepackage{algorithmicx}%
\usepackage{algpseudocode}%
\usepackage{listings}%
\usepackage{url}%
\usepackage{tabularx} %

\def\tsc#1{\csdef{#1}{\textsc{\lowercase{#1}}\xspace}}
\tsc{WGM}
\tsc{QE}
\tsc{EP}
\tsc{PMS}
\tsc{BEC}
\tsc{DE}

\begin{document}
\let\WriteBookmarks\relax
\def\floatpagepagefraction{1}
\def\textpagefraction{.001}

\shorttitle{Explainability Analysis - Large Language Model (LLM) Approach}

\shortauthors{Uddin et~al.}

\title [mode = title]{An Explainable Transformer-based Model for Phishing Email Detection: A Large Language Model Approach}                      


%
\author[1,2]{Mohammad Amaz Uddin}[
                        orcid=0000-0002-0416-5676]





\credit{Literature Review, Methodology, Software, Data curation, Implementation, Experimental analysis, Conceptualization, Writing - Original draft preparation}

\affiliation[1]{organization={Department of Computer Science and Engineering, BGC Trust University Bangladesh},
    city={Chittagong},
    postcode={4381}, 
    country={Bangladesh}}

\author[2]{Md Mahiuddin}[style=chinese]


\credit{Writing - Review \& editing and Supervision}

\affiliation[2]{organization={Department of Computer Science and Engineering, International Islamic University Chittagong},
    city={Chittagong},
    postcode={4318}, 
    country={Bangladesh}}

\author[3]{Iqbal H. Sarker}[style=chinese]

\cormark[1]
\ead{m.sarker@ecu.edu.au}
\ead[URL]{https://orcid.org/0000-0003-1740-5517}

\credit{Conceptualization, Writing - Review \& editing and Supervision}

\affiliation[3]{organization={Centre for Securing Digital Futures, School of Science, Edith Cowan University},
    city={Perth},
    postcode={WA-6027}, 
    country={Australia}}

\cortext[cor1]{Corresponding author}


\begin{abstract}
Phishing email is a serious cyber threat that tries to deceive users by sending false emails with the intention of stealing confidential information or causing financial harm. Attackers, often posing as trustworthy entities, exploit technological advancements and sophistication to make the detection and prevention of phishing more challenging. Despite extensive academic research, phishing detection remains an ongoing and formidable challenge in the cybersecurity landscape. Large Language Models (LLMs) and Masked Language Models (MLMs) possess immense potential to offer innovative solutions to address long-standing challenges. In this research paper, we present a fine-tuned transformer-based language model, RoBERTa (Robustly Optimized BERT Pretraining Approach), for detecting phishing emails. In the detection process, we employ a real-world phishing email dataset and apply the preprocessing techniques to clean and address the class imbalance issues, thereby enhancing model performance. Our experimental results demonstrate that the fine-tuned model achieves high accuracy, confirming its robustness and effectiveness in the phishing email classification task. To ensure model transparency and trust in the model’s decision-making, we propose a hybrid explanation approach, LITA (LIME-Transformer Attribution), which integrates the potential of Local Interpretable Model-Agnostic Explanations (LIME) and Transformers Interpret methods. This hybrid method provides more user-friendly insights and solves the disagreement issues between the two approaches. Moreover, we also present how the model generates its predictions by presenting positive and negative contribution scores using LIME, Transformers Interpret, and LITA.
\end{abstract}



\maketitle

\section{Introduction}

The rapid advancement of internet technology has significantly changed the way people communicate online while also raising more security issues \citep{salloum2021phishing}. One of the prevalent security concerns, the cyberattack termed “email phishing” poses grave risks such as identity or login credentials theft, financial repercussions, and harm to an organization’s reputation \citep{almomani2013survey}. In this current decade, email phishing has changed throughout time to become a more effective method for carrying out fraudulent actions. As phishers continuously innovate new techniques, detecting phishing has become increasingly challenging, posing a formidable task for law enforcement agencies, governments, and organizations combating this cyber threat. To evade falling victim to this threat, organizations need to adeptly detect and prevent these attacks at scale by adopting effective phishing detection techniques. Leveraging Artificial Intelligence (AI) techniques like Machine Learning, Deep Learning, and Hybrid Learning, which are already utilized for phishing attack detection, can significantly bolster these efforts \citep{basit2021comprehensive} \citep{sarker2023machine}. Moreover, because of the rapid advancements in technology, new methods based on AI are being developed daily in the cybersecurity field.

Transformer-based models have become more popular in the context of text classification. The transformer is an innovative neural design that uses the attention mechanism to encode incoming data into powerful characteristics \citep{han2021transformer}. Transformer-based models have greatly improved the processing, comprehension, and interpretation of text data inputs, which has had a revolutionary effect on the development of phishing classification techniques \citep{jamal2023improved}. LLMs like OpenAI's ChatGPT (Generative Pre-trained Transformer) and Google's Bard have transformed the landscape of natural language creation and comprehension, capturing considerable attention from the general public \citep{zhao2023survey, yao2023survey}. The increasing complexity of LLMs is radically changing artificial intelligence and has the potential to transform algorithm development and application by understanding human language through extensive text data patterns and correlations. 

Many transformer-based models used in natural language processing (NLP), including well-known ones like OpenAI's GPT and Google's BERT (Bidirectional Encoder Representations from Transformers), are categorized as language models due to their large number of parameters and are designed to process and generate human language. The BERT model, a type of transformer-based neural network architecture pre-trained on extensive text corpora to capture contextual word embeddings \citep{koroteev2021bert}, has become increasingly efficient and valuable in text classification and generation tasks, including spam and phishing detection, sentiment analysis \citep{singh2021sentiment}, and various other fields. BERT utilizes a bidirectional representation approach that considers both left and right context simultaneously at all layers, allowing it to learn rich language representations from large amounts of unlabeled text \citep{devlin2018bert}. With simply one extra output layer, this method allows the pre-trained BERT model to be refined, leading to the development of state-of-the-art models for a variety of applications. In order to model language, BERT uses a Masked Language Model, in which certain words in a sentence are randomly replaced with a unique token (usually [MASK]). Next, by examining the contextual clues offered by nearby words, the model is trained to infer the original words. BERT makes use of a Transformer architecture technique called “self-attention”, which allows it to recognize inter-word relationships in text sequences and assess the relative importance of individual words within a sentence in order to understand the context. Based on the transformer layers number and parameters BERT model has different variants, such as: BERT-base \citep{khadhraoui2022survey}, ALBERT \citep{lan2019albert}, RoBERTa \citep{liu2019roberta}, DistilBERT \citep{sanh2019distilbert}, etc. 

Despite BERT's versatility across NLP tasks, comprehending its internal mechanisms often poses challenges, resembling a “black box” due to its complexity. Unraveling how the model operates and forms predictions demands considerable effort \citep{xu2019explainable}. XAI techniques emerge as a vital solution, offering significant perceptions of the model's reasoning and illuminating the decision-making process \citep{hoffman2018metrics, anan2023interpretable}. Recently, Sarker \citep{sarker2024ai,sarker2024explainable} has also highlighted the significance of explainability analysis in various perspectives in the context of cybersecurity, in which we are interested in exploring more with experiments. Explainable methods can highlight the words, sub-words, or features that the language model considers most influential in various text classification tasks. In our previous work \citep{uddin2025explainabledetector}, we implemented two explainable methods and provided explanations with positive and negative coefficients of words. In this research, our main goal is to focus transparency in phishing email detection with explainable AI and to propose a hybrid method that considers the limitations of existing approaches. The major contributions of the work are as follows:

\begin{itemize}
\item We build a robust phishing email detection model by fine-tuning a pre-trained RoBERTa transformer-based language model.

\item We present a hybrid explanation approach, LITA, to provide a more user-friendly explanation for transformer-based phishing email detection. In addition, to produce alternate explanations, we also use LIME and  Transformers Interpret, which enables us to compare results and draw attention to the limitations of current XAI methods.

\item We calculate each word's positive and negative contribution score in a phishing email to show how it affects the phishing classification result.

\item We evaluate our method on a real-world dataset of phishing emails, showcasing the efficacy of the model and the clarity of the explanations it generates.

\end{itemize}

The remaining sections of the paper are structured as follows: Section 2 delves into related research works. Section 3 offers a comprehensive explanation of the proposed model's framework and methodology. In Section 4, the experimental outcomes and results are elaborated upon. Section 5 examines the transparency of the model using XAI. Moreover, within Section 6, there is an extensive discussion of the results. Lastly, Section 7 presents the concluding remarks and outlines the future prospects of this work.

\section{Literature Review}

The field of cybersecurity has grown increasingly important and complex in the modern world. There are several subfields within cybersecurity, including data security, application security, network security, etc. Cyber-attacks of all kinds are becoming more and more frequent, and they can destroy or steal imported data. In addition, in this cybersecurity field, there is an enormous amount of new and creative techniques being developed daily for identifying and preventing cyberattacks. For the improvement of phishing email detection and prevention systems, many researchers have developed a wide range of novel strategies to counteract phishing emails, utilizing state-of-the-art methods from deep learning, machine learning, and hybrid models. Phishing is still a serious cybersecurity concern and one of the main ways hackers breach networks.

\subsection{Classical Machine Learning Methods}\label{subsec1}
Detecting phishing attempts early on is critical for businesses nowadays. The security of an organization can be significantly increased by machine learning, which can identify potential assaults early on \citep{apruzzese2023role}. In ref. \citep{yasin2016intelligent}, a text processing, data mining, and knowledge discovery approach was combined to create an intelligent phishing email detection model. They utilized text stemming and WordNet ontology in pre-processing to enrich the model with synonyms and employed five machine learning classification algorithms (J48, Naïve Bayes, Support Vector Machine (SVM), Multi-Layer Perceptron (MLP), and Random Forest (RF)). They have notably achieved 99.1\% accuracy with RF and 98.4\% using J48 for an accredited dataset. The study \citep{harikrishnan2018machine} employed techniques such as SVD and NMF for feature extraction and dimensionality reduction from legitimate and phishing email text data. For the numeric representation of phishing emails, the author integrated TFIDF with the SVD and NMF, and Decision Trees, Logistic Regression, Random Forest, Naive Bayes, K-Nearest Neighbors (KNN), AdaBoost, and SVM have been utilized as classifiers. \cite{hamid2013using} presented a hybrid feature selection technique combining content-based and behaviour-based approaches to detect phishing emails. Identifying the behaviour-based features in phishing emails was the main objective of this research because an attacker cannot disguise these types of features.  They examined the overall process with a phishing dataset that consists of 6923 emails, where phishing emails were collected from the Nazario dataset and legitimate emails from the SpamAssassin dataset, and achieved 94\% accuracy with this hybrid feature selection technique. With an emphasis on the importance of top-ranking features, the study \citep{zamir2020phishing} examines feature selection methods (GR, IG, Relief-F, RFE) on a dataset of 11,055 phishing sites with 32 attributes. To improve classifier accuracy, it suggests new attributes that are generated from the strongest and weakest features. A variety of supervised learning methods with differing feature sets are compared, such as RF, SVM, bagging, kNN, Neural Network (NN), and J48. The proposed method using RF showed 97.3\%, and using stacking achieved 97.4\% accuracy. \cite{almomani2013phishing} presented the Phishing Dynamic Evolving Neural Fuzzy Framework (PDENF), which detects and predicts unknown "zero-day" phishing emails with reduced false positive and false negative rates. It leverages hybrid and evolving connectionist systems for rapid, lifelong learning, featuring a minimal memory footprint and a simplified rule base for email classification. \cite{salhi2021email} demonstrated a novel email clustering approach that uses information gain techniques to evaluate email attributes, enabling effective differentiation between benign and harmful emails.

\subsection{Deep Learning Methods}\label{subsec2}
In cybersecurity, deep learning has great potential, particularly for detecting network intrusions, classifying malware, and recognizing phishing attempts. \cite{alhogail2021applying} proposed a deep learning-based phishing email detection model based on graph convolutional network (GCN) and natural language processing (NLP). A publicly available balanced phishing dataset has been utilized, where the authors extracted features from the email body’s text to improve the proposed model’s accuracy. The model showed 98.2\% accuracy with a low false-positive rate of 0.015\%. For phishing detection, a new Intelligent Cuckoo Search (CS) Optimization Algorithm with a Deep Learning-based Phishing Email Detection and Classification (ICSOA-DLPEC) approach has been proposed in \citep{brindha2023intelligent}. Data preprocessing (email cleaning, tokenization, and stop-word elimination) techniques were employed to derive the important features from the Enron email dataset. The Gated Recurrent Unit (GRU) model was used with a CS algorithm for the fine-tuning of the parameters of the proposed model and achieved a high accuracy of 99.72\%. \cite{dewis2022phish} employed Long short-term memory (LSTM) model with NLP to detect both phishing and spam emails. They developed a phish responder, which is a Python-based command-line solution, and used deep learning and NLP for this classification task. The proposed LSTM model achieved 99\% accuracy for the text-based dataset. \cite{fang2019phishing} introduced THEMIS, a phishing email detection model built on an improved recurrent convolutional neural network (RCNN) model, which was improved by using the Bidirectional Long Short-Term Memory (Bi-LSTM). 99.84\% accuracy is demonstrated by the suggested THEMIS model on datasets that are not balanced, where the positive rate (FPR) is 0.043\%.

In addition, researchers investigated deep neural architectures such as Recurrent Neural Networks (RNNs) and Deep Belief Networks (DBNs) \citep{zhang2017phishing, bahnsen2017classifying} in the context of phishing detection, as well as a Neural Network (NN) based on Reinforcement Learning (RL), called DENNuRL, for adaptive decision-making \citep{smadi2018detection}.

\subsection{Transformer-based Methods}\label{subsec1}
Moreover, transformer-based models are also making a good impact in the cybersecurity field.\cite{thapa2023evaluation} concentrated on two specific models: RNN and BERT to test how well these models could identify phishing emails. They looked at how well these models performed when the email data was distributed unevenly across several organizations and when added more data. In this experiment, the BERT model's test accuracy was 96.1\%. In addition, another study \citep{atawneh2023phishing} focused on different deep learning architectures (CNNs, LSTMs, RNNs, BERT) to identify the emails as legitimate or phishing by analyzing different features of emails. The proposed model using BERT and LSTM, reaching an impressive accuracy of 99.61\% on publicly available 4 different datasets, outperforming other methods. Nowadays, other variants of the BERT model, such as DistilBERT, RoBERTa, and TinyBERT \citep{jiao2019tinybert}, are performing well in the phishing and spam detection research sector \citep{jamal2023improved, lee2020catbert, songailaite2023bert}. The authors \citep{jamal2023improved} presented an Improved Phishing and Spam Detection Model (IPSDM) which consists of two BERT variants, such as the fine-tuned DistilBERT and RoBERTA model,s and showed good accuracy in this research scope. Lee et al. \cite{lee2020catbert} proposed Context-Aware Tiny Bert (CatBERT), which exhibits faster processing and robustness to adversarial attacks than other transformer models, especially those that involve intentional keyword substitutions with synonyms or typos. The detection rate of this proposed model was 87\%, which was far better than other compared models. Furthermore, researchers also used fine-tuned BERT-based models for phishing URL detection \citep{wang2023large, maneriker2021urltran}. 

Overall, this literature survey shows that although the existing studies demonstrate good performance using different datasets and models, most of them do not address model interpretability or explainability. Questions such as “How does the model work?” or “Which features contribute to phishing detection?” can arise in their methodologies. In this work, to answer those types of questions and solve the existing drawbacks in the literature, we have worked with explainable AI using not only one technique but also used two different techniques with different methodologies. Although several studies have conducted comprehensive experiments on phishing email detection, many of them relied on spam datasets rather than datasets specifically designed for phishing, which may limit the validity of their findings. In cybersecurity, we know that two different types of emails exist: spam and phishing, and not all spam emails are considered to be phishing. For this reason, we only focused on the phishing email dataset.

\section{Methodology}

\begin{figure}[htbp]
\centering
 \includegraphics[width=0.85\linewidth]{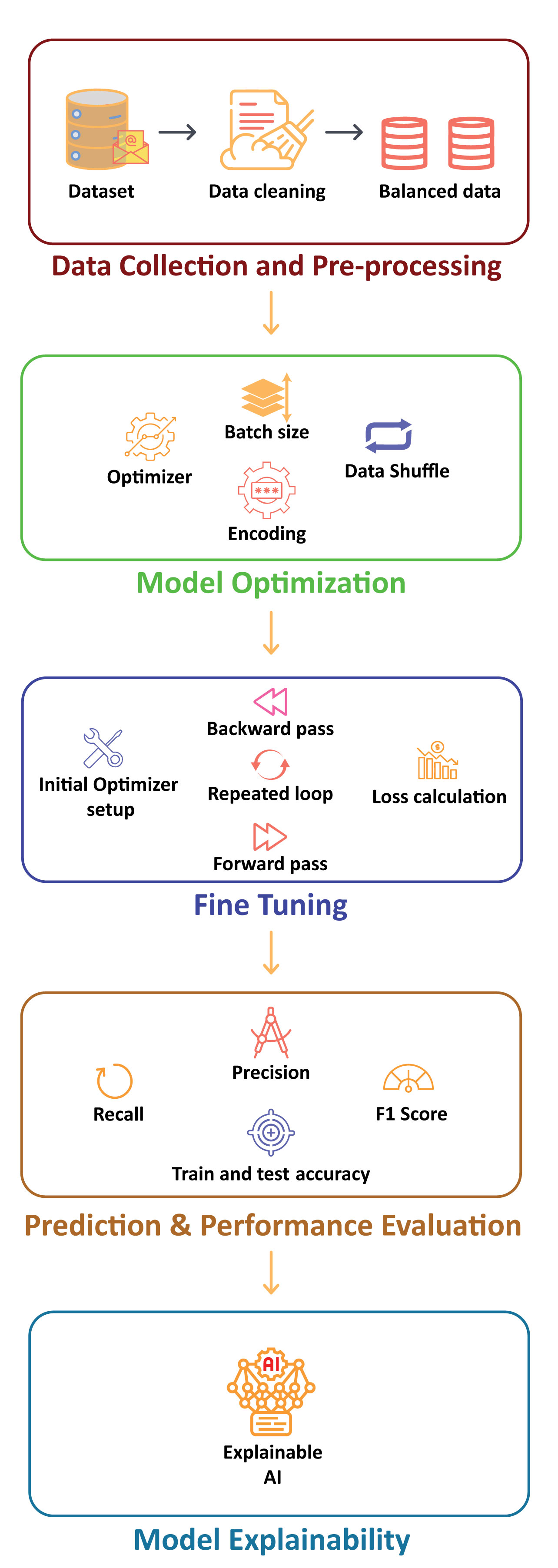}
  \caption{An overview of our experimental methodology.}
  \label{fig:Methodology}
\end{figure}

In this paper, we have tried to explore the Transformer-based self-attention mechanism BERT model with explainable AI. Initially, we collected a dataset and prepared it with data cleaning and balancing techniques. After preparing the data, we have focused on optimizing and fine-tuning the model. For this phishing email detection purpose, we have selected the RoBERTa model based on our previous experimental analysis \cite{uddin2025explainabledetector} and pre-trained it with the prepared data. The RoBERTa model demonstrates good performance in both balanced and imbalanced scenarios of datasets when identifying phishing emails from texts. At the end of the experiment, we employed XAI techniques to explore our fine-tuned model and described how it is working and performing with good accuracy in this text classification. The methodology of the proposed work is depicted in Fig.~\ref{fig:Methodology}.

\subsection{Data Characteristics and Processing}\label{subsec1}
\begin{figure*}[htbp]
    \centering
    \includegraphics[width=\linewidth]{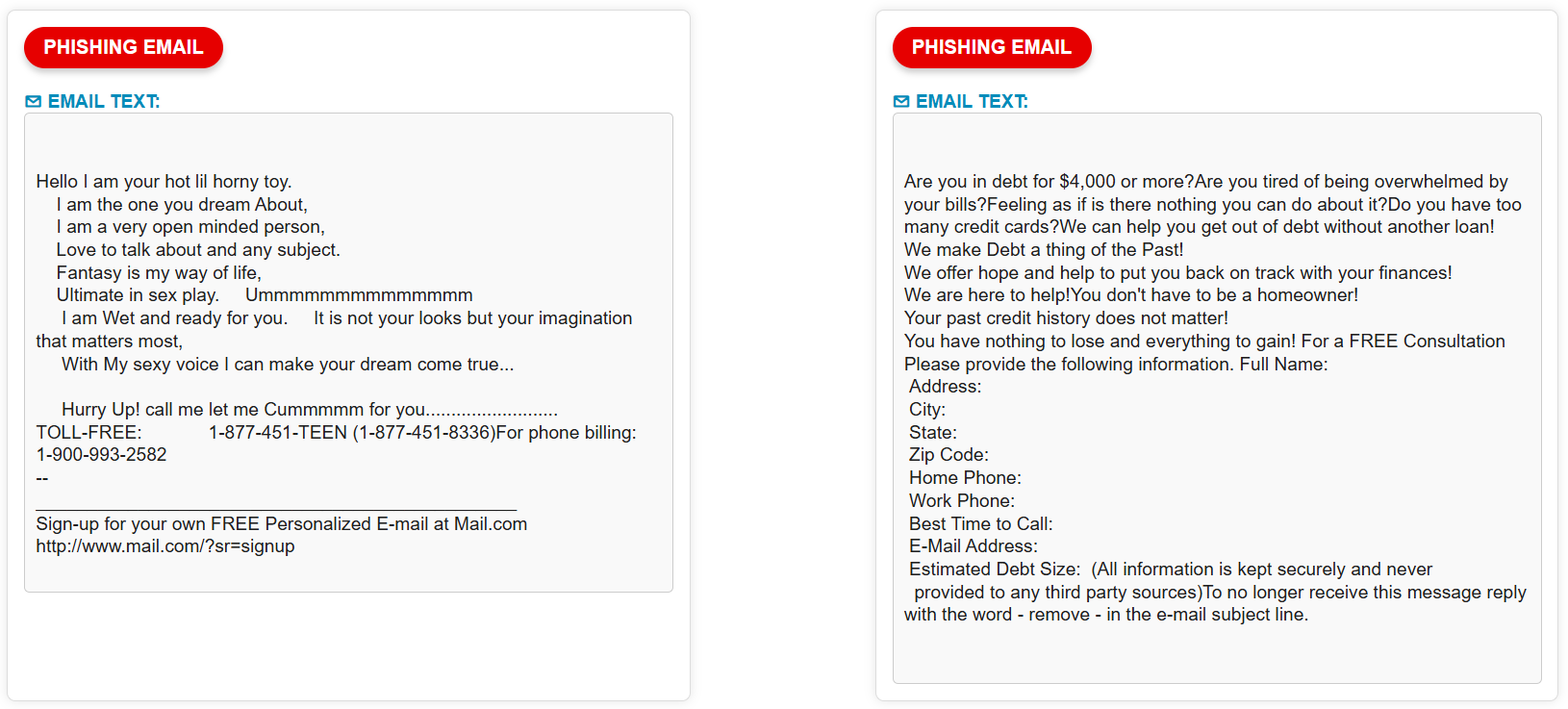}
    \caption{The sample data of Phishing Emails}
  \label{fig:Data}
\end{figure*}

For this experiment, a publicly available dataset has been collected from Kaggle. The dataset describes different kinds of email text bodies that can be used to identify phishing emails using in-depth text analysis. This dataset consists of two different types of email: Legitimate or Legitimate Email and Phishing Email. Fig.~\ref{fig:Data} shows the phishing email samples. The dataset comprises 11,322 legitimate Emails and 7,328 phishing Emails, indicating that it is imbalanced. Additionally, this dataset revealed the presence of null values in certain rows. As null values can’t be processed, that’s why those rows are removed. 

We initially focused on data cleaning before addressing the data imbalance issue. This process involved converting the input text to lowercase, removing HTML tags that may be present in the text, replacing multiple whitespace characters with a single space, removing carriage return characters, and trimming leading and trailing whitespace. These steps ensured that the text was in a clear and consistent format for further processing.

After cleaning the data, we utilized a data augmentation technique called synonym replacement to address the imbalance problem. This technique replaces random words from the original text with their synonyms, preserving the original meaning of the sentence while maintaining its similarity to the original text. In this context, we randomly selected words that are not common stopwords and replaced them with their randomly chosen synonyms. 

\begin{figure*}[htbp]
    \centering
    \begin{subfigure}{0.4\textwidth}
        \includegraphics[width=\linewidth]{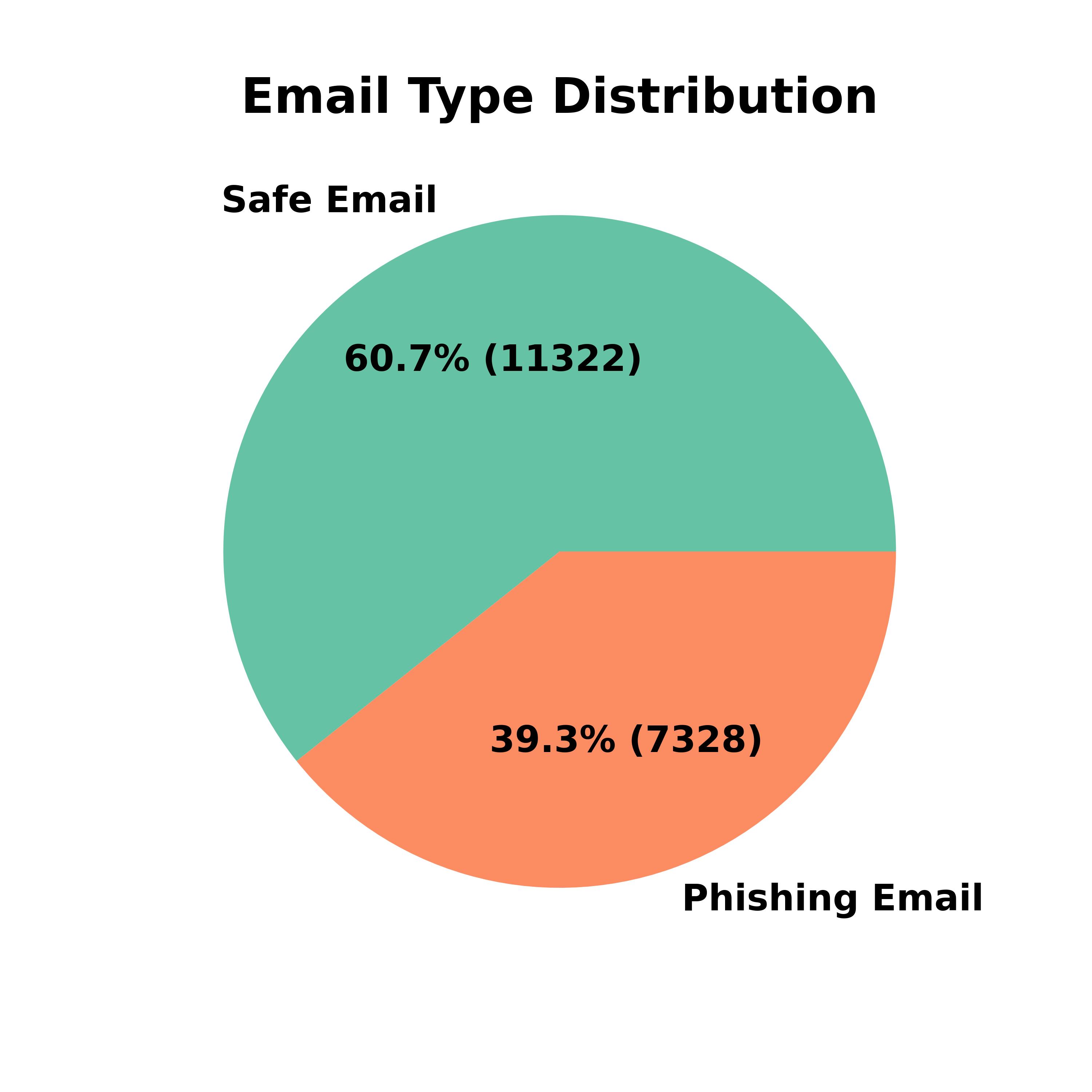}
        \caption{Imbalanced data}
    \end{subfigure}%
    \begin{subfigure}{0.4\textwidth}
        \includegraphics[width=\linewidth]{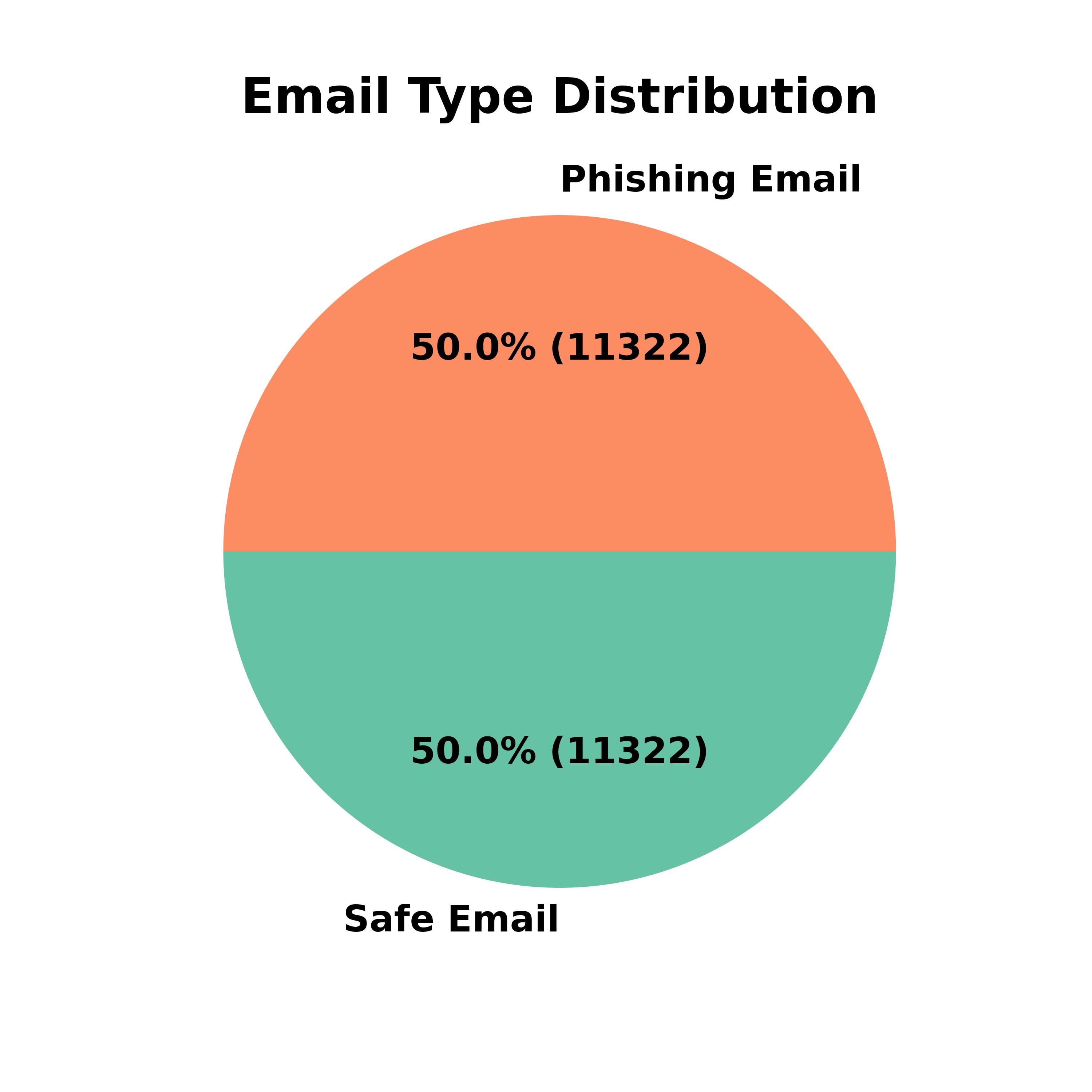}
        \caption{Balanced data}
    \end{subfigure}
    \caption{Imbalanced and balanced data.}
    \label{fig:3}
\end{figure*}

By using the augmentation technique, the minor class “Phishing Email” is oversampled. It is essential to address uneven data to develop equitable and precise predictive models. Balancing data ensures algorithms work well across all categories by reducing biases toward the majority class. This prevents skewed outputs and makes it possible to create a more accurate and equitable model for underrepresented classes. After oversampling, the dataset has 11,322 legitimate emails and 11,322 phishing emails in equal proportion. This data balancing also helps to reduce the risk of oversampling. Fig.~\ref{fig:3} illustrates the imbalanced and balanced data before and after oversampling.

\subsection{Dataset splitting}\label{subsec2}
For this experiment, we split the dataset into 80-20\%, where 80\% of the data is used for training and 20\% for testing. The 80\% training data is then further split into 80-20\%, resulting in an overall split of 64\% training set, 16\% validation set, and 20\% testing set. The 64\% training set contains a total of 14,492 data, the validation set consists of 3,623 data, and the testing set comprises 4,529 data. Initially, the training set was used to train the model, the validation set was used to validate the model to ensure its generalizability, and the testing set was used to evaluate the model's effectiveness in identifying and predicting phishing on unseen data.

\subsection{Model Selection}\label{subsec3}
\begin{figure}[h]
  \centering
  \includegraphics[width=\columnwidth]{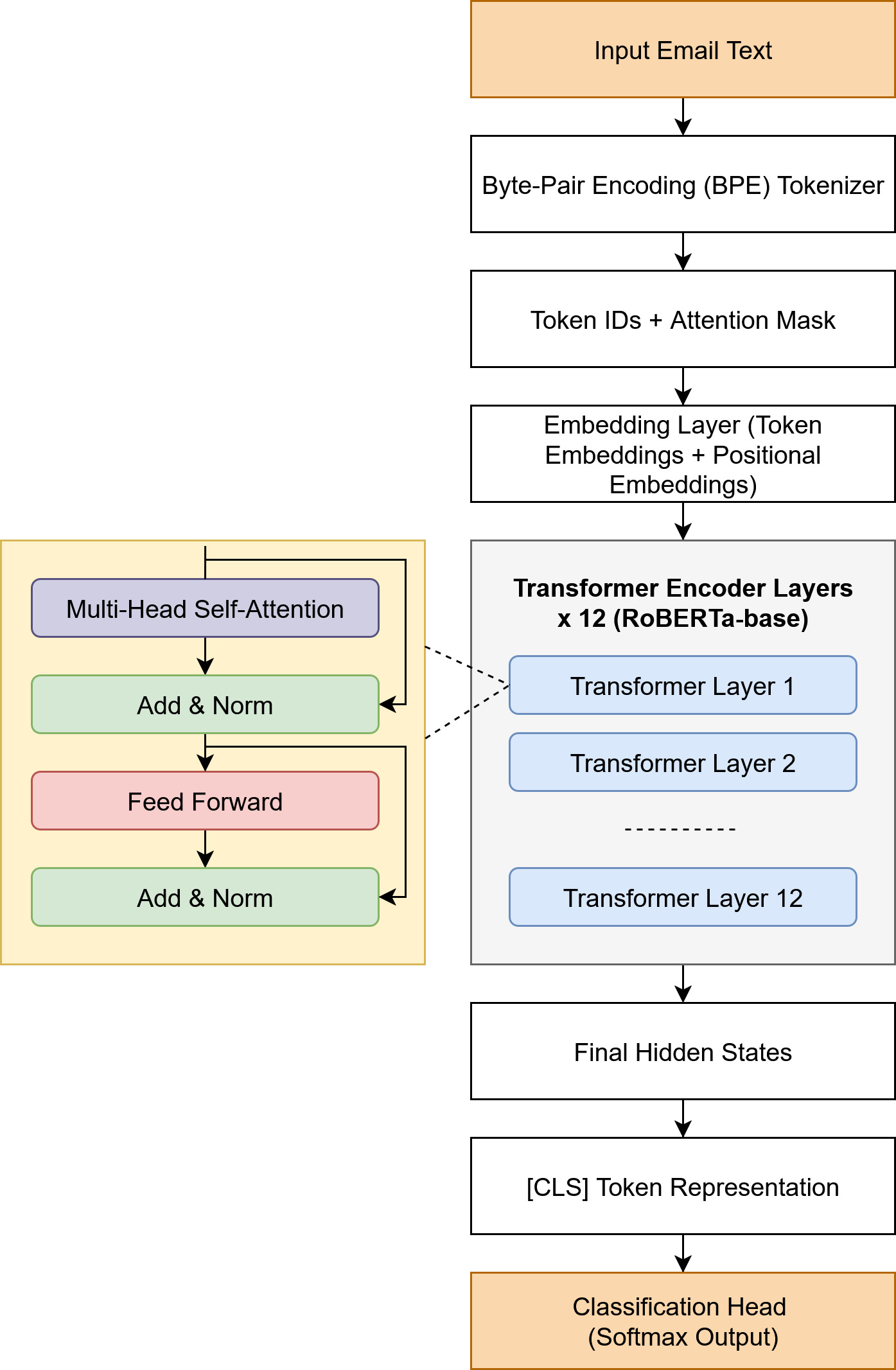}
  \caption{The architecture of the RoBERTa model}
  \label{fig:model_architecture}
\end{figure}

In this experiment, we have employed RoBERTa, a transformer-based robustly optimized variant of BERT, which retains BERT's architecture and has refined the training process through key modifications. RoBERTa is trained on a much larger text dataset (160 GB of text) using 500,000 training steps with a large batch size of 8,000 per unit\citep{liu2019roberta}. The training process benefited greatly from dynamic masking, which updates the masking scheme during training instead of using static masking like BERT did during data preprocessing. This allows the model to learn a much more diverse set of training examples (from the same text) and also improves robustness. Moreover, the model removes the Next Sentence Prediction (NSP) task that BERT used, only uses MLM, finding it unnecessary and potentially less helpful to performance. Additionally, it uses a more efficient tokenization method, Byte-Level Byte Pair Encoding (BPE), similar to GPT-2, which helps handle rare words better. The RoBERTa-base contains about 125 million parameters. This variant has 12 attention heads, a 768-dimensional hidden state, and 12 transformer layers. The architecture of the RoBERTa model is illustrated in Fig. ~\ref{fig:model_architecture}.

\subsubsection{Model Optimization}\label{subsubsec1}
RoBERTa employs a BPE tokenizer that is capable of splitting words into smaller sub-word units to better manage out-of-vocabulary terms and to allow for more meaningful sub-word representations. The preprocessed phishing dataset is tokenized using RoBERTaTokenizer, the default tokenizer for the RoBERTa transformer model. The maximum sequence length of the input data is configured to be a maximum of 512 tokens during the tokenization process, as this is the standard length for most transformer-based architectures, including RoBERTa. Moreover, as we are working with an email dataset, where the email texts are often long, setting the token length to less than 512 could result in a potential loss of important information. Other important steps in the tokenization process include padding and truncation: padding fills in sequences that are shorter than the maximum length with special token padding tokens, and truncation removes tokens from the end of sequences that are longer than the maximum length. These two tasks are important as they ensure input dimensions are standardized across the algorithm, so that embeddings are consistent and the batches of data can be processed efficiently during training and inference time. 

Pre-trained weights acquired throughout the pre-training phase are used to initialize the RoBERTa model. Selecting an optimal batch size is paramount in deep learning, as it significantly influences both training and testing accuracies, along with overall runtime \citep{lin2022analysis}. The careful selection of batch size determines the efficiency of the neural network training process, striking a balance between computational resources and model performance. The batch sizes for training and testing data are optimized at 32 and 64, respectively. Each epoch's training data is assigned at random to ensure the model sees a diversity of unseen data samples and prevents overfitting. This shuffling method increases the model's ability to generalize.

During training, an optimizer is essential in adjusting the parameters of a neural network. The main goal of this is to minimize the error or loss function of the model for better performance. For this research work, AdamW (Adam Weight Decay) \citep{loshchilov2017decoupled} optimizer has been used to get better performance and avoid overfitting. This optimizer is a modified version of Adam, which is widely adopted for training deep neural networks \citep{zhuang2022understanding}. By separating weight decay from the learning rate, AdamW enhances its implementation. It improves regularization and training dynamics by separating weight decay from gradient updates, in contrast to Adam. Learning rate is a crucial hyperparameter governing the step size during training for optimizing models. The convergence and overall performance of the system are dependent on various aspects such as the model architecture, optimization techniques, and task domain. For efficient fine-tuning and fruitful model training, this parameter must be adjusted. Selecting a high learning rate can introduce instability, resulting in poor performance on unseen data, while a low learning rate may extend the training process, requiring more epochs and increasing computational costs. In this research, a learning rate, 2e-5 is used in the training process for the transformer-based RoBERTa model.

The overall discussion about model optimization indicates the parameter settings for the introduced best model for this specific experiment. The hyperparameter settings for the suggested model are shown in Table~\ref{tbl1}. We worked with every possible value of the parameters by using an iterative process to find the ideal value. Following that, these improved hyperparameter settings are used to train the suggested model.\\

\begin{table}[htbp]
\caption{Parameter Settings for Phishing Email Detection Task.}\label{tbl1}
\begin{tabularx}{\linewidth}{@{} llc @{}}
\toprule
Parameter Name & Worked With & Optimal Value\\
\midrule
Training Batch size & 8, 16, 32 & 32 \\
Testing Batch size & 8, 16, 32, 64 & 64 \\
Optimizer & AdamW & AdamW \\
Learning rate & 1e-5, 2e-5, 3e-5 & 2e-5 \\
Number of epochs & 3-10 & 5 \\
\bottomrule
\end{tabularx}
\end{table}

\subsubsection{Fine Tuning}\label{subsubsec2}
\begin{figure}[htbp]
    \centering
    \includegraphics[width=\linewidth]{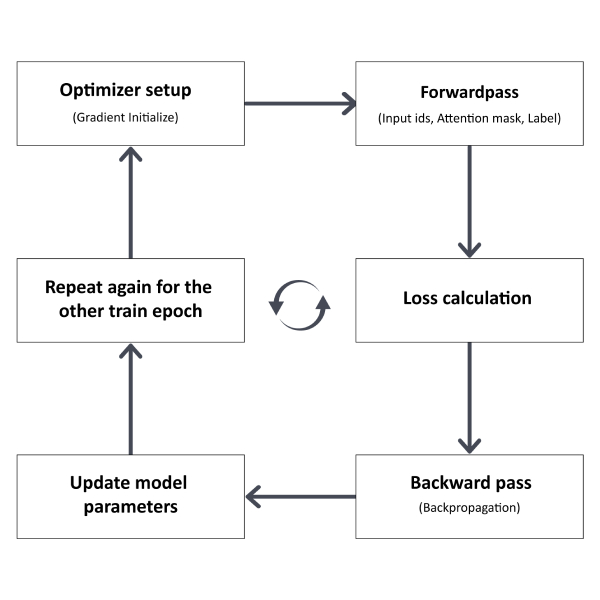}
    \caption{The process of fine-tuning.}
    \label{fig:Fine_Tuning}
\end{figure}

A pre-trained model is fine-tuned using task-specific data and backpropagation to alter parameters on certain tasks and datasets. The primary objective of this is to improve the model's performance across various tasks and datasets while minimizing the loss function. In this experiment, RobertaForSequenceClassification has been utilized to adopt the pre-trained model. DataLoader is used to configure the train\_loader and test\_loader by setting prepared data, batch size, and shuffling command, where train\_loader is configured with shuffling and test\_loader is configured without shuffling. After that, we fine-tuned this RoBERTa model during the training process by adjusting and adopting some necessary steps: optimizer setup, forward pass, loss calculation, backward pass (backpropagation), and updating model parameters, which are shown in Fig.~\ref {fig:Fine_Tuning}.

In this overall fine-tuning process, following the optimizer's setup, the model is fine-tuned by feeding it the input parameters (input\_ids, attention\_mask, and label) during the forward pass so that predictions can be made. After the forward pass, gradients of the loss for the parameters (weights and biases) are calculated in the backward pass. These gradients show how changing the parameters would affect the loss. Lastly, an update is made to the model parameters. The optimizer is mainly responsible for changing the neural network's parameters in accordance with the calculated gradients. It adjusts the weights and biases in an attempt to reduce the loss. The model gains the capacity to fine-tune its parameters throughout several epochs, which improves its predictive capability on the training set. The ultimate goal is to identify underlying patterns in the training set and use those patterns to generalize well to new data. 

\subsubsection{Model Explainability}\label{subsubsec3}

\begin{figure*}
\centering
\includegraphics[width=0.95\textwidth, height=0.95\textheight, keepaspectratio]{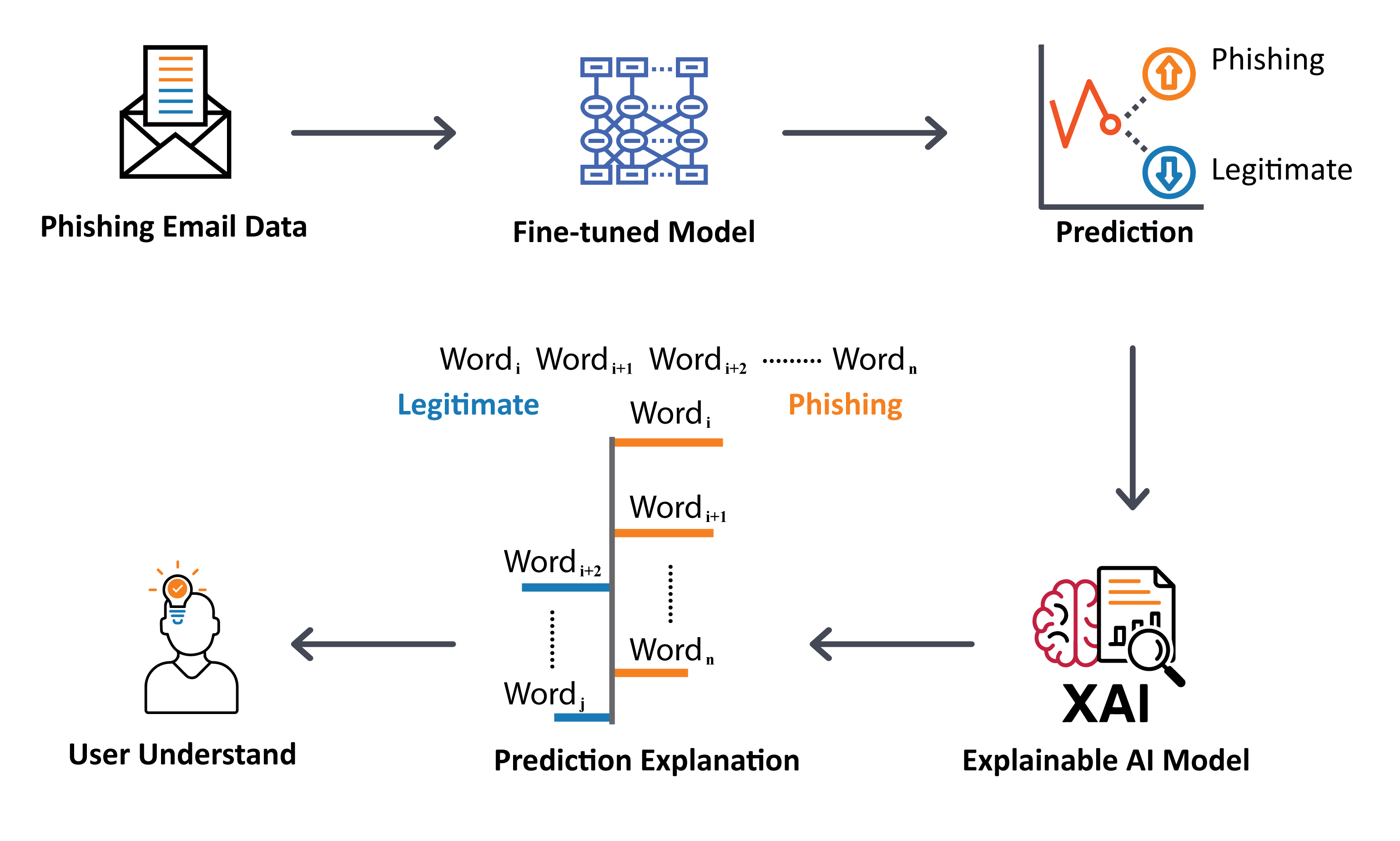}
  \caption{The overall process of model explainability}
  \label{fig:Model_Explainability}
\end{figure*}

The ability of a model to explain its decisions is crucial since it provides insight into the process. Verifying the accuracy of judgments and reducing errors are made feasible by understanding how models work. XAI has been a popular area with a growing group that has developed efficient techniques to understand deep neural network predictions among the more complex machine learning models \citep{holzinger2022explainable}. This is important because sophisticated models frequently act as opaque “black boxes" making it difficult for people to comprehend the logic behind particular decisions \cite{sarker2024ai}. In this research work, we have worked with model explainability to solve the black box problem. To understand the model prediction, two popular explainability models, LIME and Transformers Interpret, have been utilized. The overall process of model explainability is shown in Fig.~\ref{fig:Model_Explainability}.

Generally, this experiment for phishing email detection is a text classification task that LimeTextExplainer used to explain the black box RoBERTa model. The purpose of the LIME method is to faithfully explain any classifier or regressor's predictions \citep{ribeiro2016should}. For improved interpretability, LIME provides an understandable depiction of the decision-making process, providing light on the “why" behind the predictions or suggestions made by these models. To achieve model explainability, LIME first perturbs the email content by making modest adjustments, such as rearranging words or adding new ones. After that, the black-box transformer-based model is applied to the altered email content to generate predictions. From these impacted instances, LIME derives features. This local model should roughly represent the behaviour of the black-box model surrounding the chosen instance. Next, explanations for the black-box model's prediction for the original data are produced using the local model. This explanation usually highlights the words or key elements most important during the model's decision-making process.

A Transformers Interpret tool was also utilized to explain the transformer-based model for this classification work.  The class explainer process was carried out using this tool, using a model and tokenizer. With the use of this class explanation, word attributions are computed. Positive attribution numbers signify a word's positive contribution to the projected class, while negative numbers signify a word's negative contribution. The terms that demonstrate a strong contribution to decision-making are also highlighted in this explanation. 

However, while LIME and Transformers Interpret both provide a suitable understanding of a model's decision-making processes, they differ fundamentally in their methods for explanation. LIME operates on word-level perturbations, whereas Transformers Interpret uses gradient-based attribution on subword tokens derived from the tokenizer. As a result, the outputs of these techniques, such as the number of highlighted tokens, the magnitude of scores, and their interpretive granularity, often vary, potentially leading to confusion for users. To solve this limitation and enhance the clarity and consistency of the explanations, we present a hybrid explanation approach called LITA. This method integrates the attribution scores from both LIME and Transformers Interpret into a unified explanation by computing a weighted combination of their normalized outputs. Algorithm ~\ref{algo1} shows the explanation process of the model prediction using a hybrid approach to generate comprehensive explanatory scores. Here, $\alpha$ is a tunable parameter in the range $[0, 1]$, controlling the influence of each technique. Both LIME (norm\_lime\_scores) and Transformers (norm\_transformer\_scores) attribution scores are first normalized to a common range $[-1, 1]$ to ensure consistency and comparability. As we are not sure which method is more accurate so we set $\alpha = 0.5$ in this experiment, which assigns equal weight to both methods. In addition to reducing the overall differences between both attribution techniques, offering an indication of total feature importance with a hybrid attribution technique offers an additional layer of interpretability and reliability of the decision-making process.

\begin{algorithm}[h]
\caption{Hybrid Explainability Method}\label{algo1}
\begin{algorithmic}[1]
\Require Trained model, tokenizer, and input text
\Ensure Explanation of the model prediction
\State class\_names $\Leftarrow$ [X, Y, ..., N]

\Function{Explain\_Prediction}{model, tokenizer, text}
    \State lime\_scores $\Leftarrow$ \Call{LIME\_Explanation}{model, text}
    \State transformer\_scores $\Leftarrow$ \Call{Transformer\_Interpret}{model, tokenizer, text}
    \State normalized\_lime $\Leftarrow$ \text{lime\_scores scaled to } [-1, 1]
    \State normalized\_transformer $\Leftarrow$ \text{transformer\_scores scaled to } [-1, 1]
    \State hybrid\_scores $\Leftarrow$ \Call{Hybrid\_Explanation}{normalized\_lime, normalized\_transformer, $\alpha$}
    \State \Return lime\_scores, transformer\_scores, hybrid\_scores
\EndFunction

\Function{LIME\_Explanation}{model, text}
    \State explainer $\Leftarrow$ LimeTextExplainer(class\_names $\Leftarrow$ class\_names)
    \State explanation $\Leftarrow$ explainer.explain\_instance(text, model, num\_features $\Leftarrow n$, top\_labels $\Leftarrow 1$)
    \State exp\_class $\Leftarrow$ explanation.available\_labels()
    \State lime\_scores $\Leftarrow$ explanation.as\_list(label $\Leftarrow$ exp\_class)
    \State \Return lime\_scores
\EndFunction

\Function{Transformer\_Interpret}{model, tokenizer, text}
    \State cls\_explainer $\Leftarrow$ SequenceClassificationExplainer(model, tokenizer)
    \State transformer\_scores $\Leftarrow$ cls\_explainer(text)
    \State \Return transformer\_scores
\EndFunction

\Function{Hybrid\_Explanation}{norm\_lime\_scores, norm\_transformer\_scores, $\alpha$}
    \ForAll{word in norm\_lime\_scores $\cap$ norm\_transformer\_scores}
        \State hybrid[word] $\Leftarrow \alpha \cdot$ norm\_lime\_scores[word] $+ (1 - \alpha) \cdot$ norm\_transformer\_scores[word]
    \EndFor
    \State \Return hybrid
\EndFunction
\end{algorithmic}
\end{algorithm}

\section{Results}

A model's generalization power can be assessed using various performance metrics, including precision, recall, F1-score, and accuracy, across different conditions \citep{chen2024hybrid}. The fine-tuned transformer-based model was trained and tested with unbalanced and balanced datasets. To assess the performance of the model, different measurements such as the confusion matrix, precision, recall, F1-score, and overall accuracy have been calculated. Overall summarize the performance, a model needs to calculate the confusion matrix first. This kind of table provides a detailed breakdown of the actual and predicted classification for specific task domain data. The RoBERTa model's confusion matrix for imbalanced and balanced datasets is shown in Table~\ref{tbl2}. True Positive (TP), True Negative (TN), False Positive (FP), and False Negative (FN) are the main components of the confusion matrix, and as well as are also needed for the calculation of Precision, recall, and F1-score \citep{uddin2022cyber}. The precision, recall, and F1-score of the RoBERTa model for both imbalanced and balanced datasets are shown in Table~\ref{tbl3} and Table~\ref{tbl4}.

\begin{table*}[htbp]
\caption{Confusion matrix of RoBERTa for both imbalanced and balanced datasets}\label{tbl2}
\begin{tabular*}{\textwidth}{@{\extracolsep{\fill}}lcccc}
\toprule
& \multicolumn{2}{c}{Imbalanced} & \multicolumn{2}{c}{Balanced} \\
\cmidrule{2-3}\cmidrule{4-5}
Class Name & Legitimate Email & Phishing Email & Legitimate Email & Phishing Email \\
\midrule
Legitimate Email  & 2150 & 59 & 2181 & 60 \\
Phishing Email  & 17  & 1501  & 10 & 2278\\
\bottomrule
\end{tabular*}
\end{table*}

\begin{table}[width=\linewidth,cols=5,pos=h]
\caption{The precision, recall, and f1-score for RoBERTa model on  imbalanced dataset.}\label{tbl3}
\begin{tabular*}{\tblwidth}{@{} LLLLL@{} }
\toprule
Class & Precision  & Recall & F1-score & Support\\
\midrule
Legitimate Email    & 0.99   & 0.97  & 0.98 & 2209  \\
Phishing Email    & 0.96   & 0.99  & 0.98 & 1518  \\
\bottomrule
\end{tabular*}
\end{table}

\begin{table}[width=\linewidth,cols=5,pos=h]
\caption{The precision, recall, and f1-score for RoBERTa model on balanced dataset.}\label{tbl4}
\begin{tabular*}{\tblwidth}{@{} LLLLL@{} }
\toprule
Class & Precision  & Recall & F1-score & Support\\
\midrule
Legitimate Email    & 1.00   & 0.97  & 0.98 & 2241  \\
Phishing Email    & 0.97   & 1.00  & 0.98 & 2288  \\
\bottomrule
\end{tabular*}
\end{table}

Precision: Precision, which expresses the ratio of true positives to all predicted positives, indicates how accurate positive predictions are and emphasizes the model's capacity to identify positive samples. 
\begin{equation}
\begin{aligned}
\text{Precision} &= \frac{TP}{TP + FP}
\end{aligned}
\end{equation}

Recall: It calculates the ratio of true positives to all predicted positives to assess a model's accuracy in identifying positive samples. It denotes the sensitivity of the model, capturing the actual positive rate in the class observations.

\begin{equation}
\begin{aligned}
\text{Recall} &= \frac{TP}{TP + FN}
\end{aligned}
\end{equation}

F1-score: A balanced statistical metric that assesses a model's efficacy in predicting positive instances while accounting for imbalanced datasets is the F1-score, which is derived from the harmonic mean of accuracy and recall. It offers a thorough perspective that balances recall and precision, impacting the final score determined by the model's projections of the positive rate and actual positive cases.

\begin{equation}
\begin{aligned}
\text{F1-score} &= 2 \times \frac{\text{Precision} \times \text{Recall}}{\text{Precision} + \text{Recall}}
\end{aligned}
\end{equation}

The frequency of valid predictions in a model is quantified by accuracy, which also represents the percentage of correctly predicted samples out of all the samples the model generates. 

\begin{equation}
\text{Accuracy} = \frac{\text{TN} + \text{TP}}{\text{TN} + \text{TP} + \text{FN} + \text{FP}}
\end{equation}

\begin{figure*}
\centering
\includegraphics[width=0.8\textwidth]{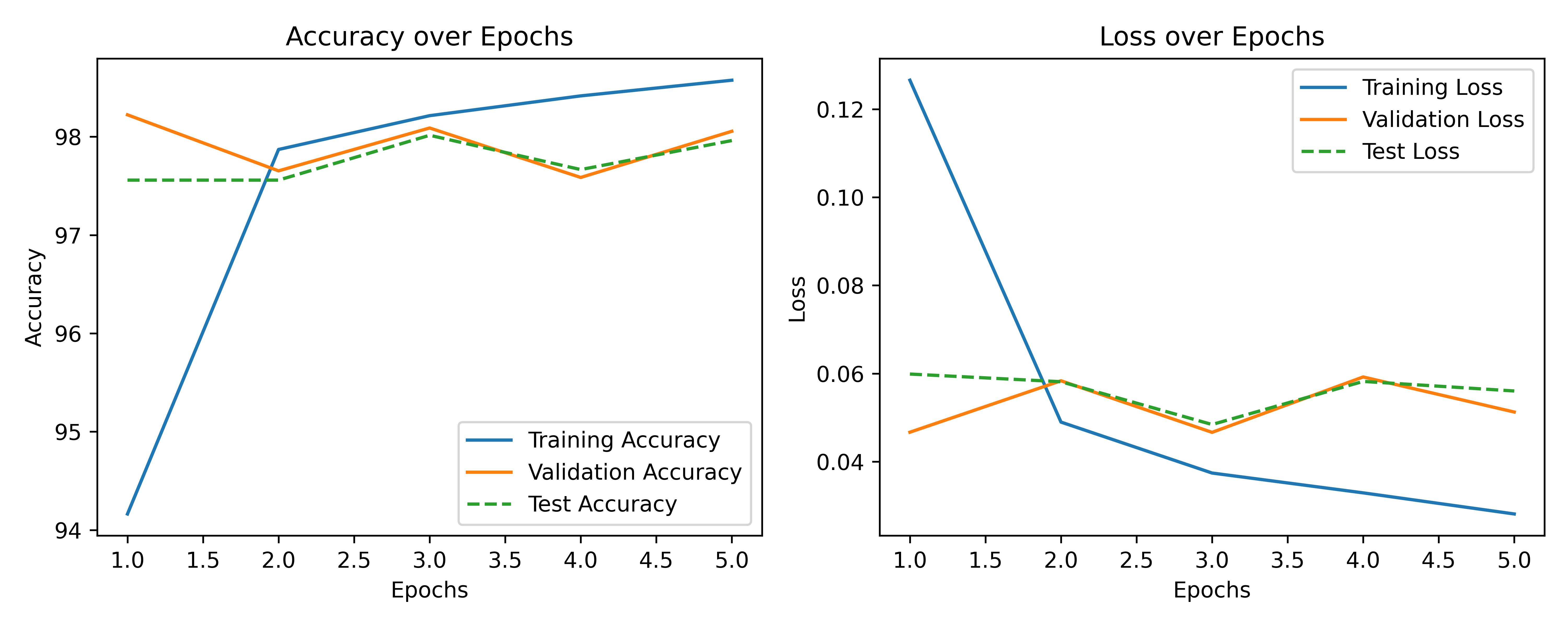}
  \caption{Training vs. Validation vs. Testing Accuracy and Loss on Imbalanced Data}
  \label{fig:Imbalanced_loss_and_accuracy}
\end{figure*}

The accuracy of a model can differ based on the various parameters, data is balanced or not. Initially, in this experiment, an imbalanced dataset was utilized to assess the model’s performance in training, validation, and testing. This dataset is highly imbalanced based on the legitimate email and phishing email data, where the number of data of legitimate emails is much higher than phishing emails. To train the model with 64\% imbalanced data, we have used 5 epochs with a batch size of 32. After 5 epochs, the model showed 98.57\% training accuracy for this imbalanced dataset with 0.0281 training loss. Then the trained model was validated with 16\% unseen data and achieved 98.05\% validation accuracy with  0.0513 loss. On the other hand, we got a testing accuracy 97.96 with testing loss 0.0560 with batch size 64. Fig. ~\ref{fig:Imbalanced_loss_and_accuracy} shows a graph of accuracy and loss with 6 epochs for an imbalanced dataset.

\begin{figure*}
\centering
\includegraphics[width=0.8\textwidth]{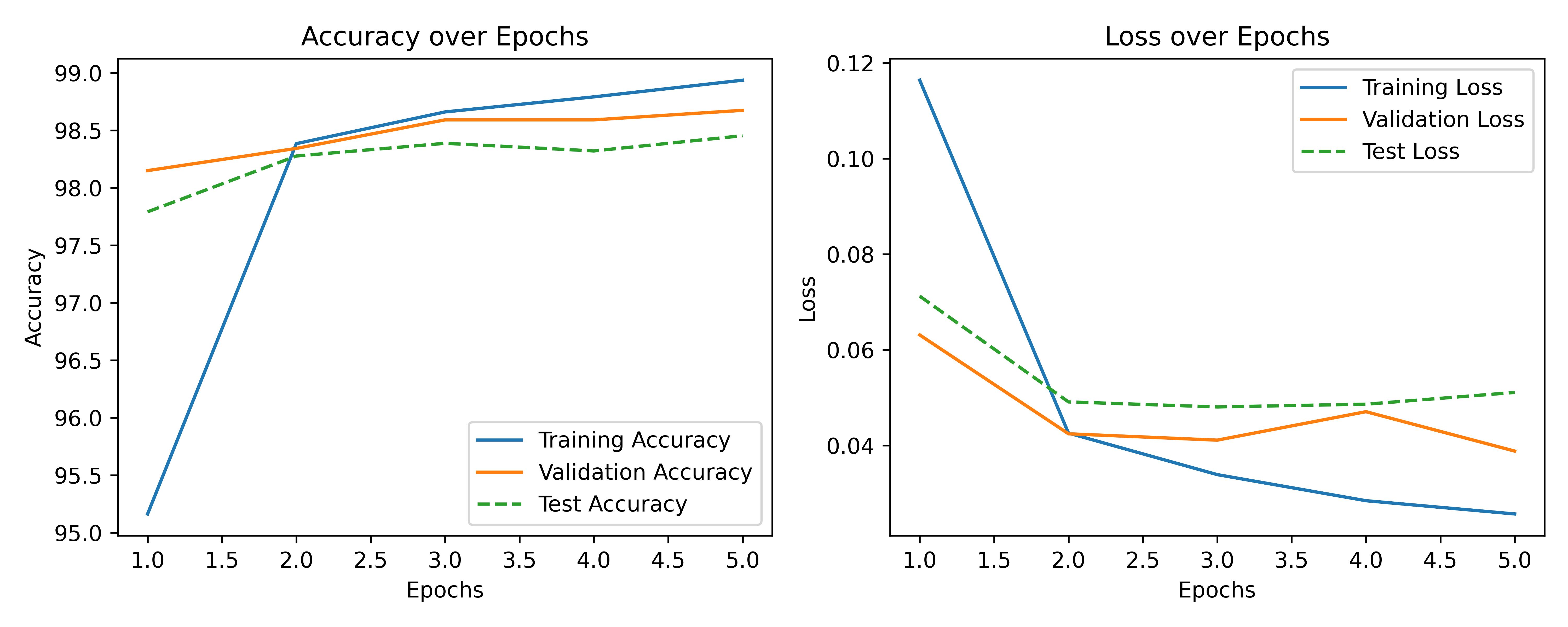}
  \caption{Training vs. Validation vs. Testing Accuracy and Loss on Balanced Data.}
  \label{fig:Balanced_loss_and_accuracy}
\end{figure*}

Moreover, we also conducted this experiment with a balanced dataset. After preparing the balanced dataset again used 64\% of the data for training purposes. We worked on model optimization and fine-tuning using this data portion and trained the model. With the balanced data, the fine-tuned model achieved 98.94\% training accuracy with 0.0257 training loss, indicating exceptional performance on the training data. After that, the fine-tuned model was validated with 16\% validation data and obtained 0.0388 validation loss with 98.68\% validation accuracy, demonstrating that the model generalizes well to unseen data and is not overfitting. Finally, the trained model was tested with 20\% unseen balanced test data and achieved 98.45\% testing accuracy with 0.0511 testing loss. This performance showed that the performance of the model with a balanced dataset is much better than an imbalanced dataset. Fig.~\ref{fig:Balanced_loss_and_accuracy} shows a graph of accuracy and loss with 5 epochs for a balanced dataset. The performance comparison of the model on both balanced and imbalanced datasets is shown in Table~\ref{tbl5}.

\begin{table}[width=\linewidth,cols=3,pos=h]
\caption{The Performance Comparison of Fine-Tuned RoBERTa Model on Imbalanced and Balanced Datasets.}\label{tbl5}
\begin{tabular*}{\tblwidth}{@{} p{3cm}p{2.3cm}p{2.2cm}@{} }
\toprule
 & \multicolumn{2}{c}{Dataset Mode} \\
\cline{2-3}
& Imbalanced & Balanced \\
\midrule
Training Loss & 0.0281 & 0.0257 \\
Training Accuracy & 98.57 & 98.94 \\
Validation Loss & 0.0513 & 0.0388 \\
Validation Accuracy & 98.05 & 98.68 \\
Testing Loss & 0.0560 & 0.0511 \\
Testing Accuracy & 97.96 & 98.45 \\
\bottomrule
\end{tabular*}
\end{table}

Additionally, we employed traditional machine learning models, including Extreme Gradient Boosting (XGB), Random Forest (RF), Support Vector Machine (SVM), and Logistic Regression (LR), to evaluate their performance and compare them with our fine-tuned RoBERTa model on an imbalanced and balanced dataset. The generalization power and robustness of the different ML models and the RoBERTa model are shown in Table~\ref{tbl6} and Table~\ref{tbl7}. Table~\ref{tbl8} and Table~\ref{tbl9} present a comparative performance analysis of these models. The overall performance analysis showed that our fine-tuned RoBERTa model outperformed the traditional ML models.

\begin{table}[width=\linewidth,cols=5,pos=h]
\caption{Precision, Recall, and F1-Score of traditional ML and RoBERTa Model for Imbalanced Dataset}
\label{tbl6}
\begin{tabular*}{\tblwidth}{@{} p{1.5cm}p{2cm}p{1cm}p{1 cm}p{1.2cm}@{} }
\toprule
Model & Class & Precision & Recall & F1-score \\
\midrule
\multirow{2}{*}{RF} & Legitimate Email & 0.96 & 0.97 & 0.96 \\
                     & Phishing Email & 0.95 & 0.94 & 0.95 \\
\midrule
\multirow{2}{*}{XGB} & Legitimate Email & 0.98 & 0.95 & 0.96 \\
                     & Phishing Email & 0.93 & 0.98 & 0.95 \\
\midrule
\multirow{2}{*}{SVM} & Legitimate Email & 0.99 & 0.97 & 0.98 \\
                     & Phishing Email & 0.95 & 0.99 & 0.97 \\
\midrule
\multirow{2}{*}{LR} & Legitimate Email & 0.98 & 0.96 & 0.97 \\
                     & Phishing Email & 0.94 & 0.97 & 0.96 \\
\midrule
\multirow{2}{*}{RoBERTa} & Legitimate Email & 0.99 & 0.97 & 0.98 \\
                            & Phishing Email & 0.96 & 0.99 & 0.98 \\
\bottomrule
\end{tabular*}
\end{table}

\begin{table}[width=\linewidth,cols=5,pos=h]
\caption{Precision, Recall, and F1-Score of traditional ML and RoBERTa Model for Balanced Dataset}
\label{tbl7}
\begin{tabular*}{\tblwidth}{@{} p{1.5cm}p{2cm}p{1cm}p{1 cm}p{1.2cm}@{} }
\toprule
Model & Class & Precision & Recall & F1-score \\
\midrule
\multirow{2}{*}{RF} & Legitimate Email & 0.98 & 0.96 & 0.97 \\
                     & Phishing Email & 0.96 & 0.98 & 0.97 \\
\midrule
\multirow{2}{*}{XGB} & Legitimate Email & 0.99 & 0.94 & 0.96 \\
                     & Phishing Email & 0.95 & 0.99 & 0.97 \\
\midrule
\multirow{2}{*}{SVM} & Legitimate Email & 0.99 & 0.97 & 0.98 \\
                     & Phishing Email & 0.97 & 0.99 & 0.98 \\
\midrule
\multirow{2}{*}{LR} & Legitimate Email & 0.99 & 0.96 & 0.97 \\
                     & Phishing Email & 0.96 & 0.99 & 0.97 \\
\midrule
\multirow{2}{*}{RoBERTa} & Legitimate Email & 1.00 & 0.97 & 0.98 \\
                            & Phishing Email & 0.97 & 1.00 & 0.98 \\
\bottomrule
\end{tabular*}
\end{table}

\begin{table}[width=\linewidth,cols=5,pos=h]
\caption{Comparative analysis of the fine-tuned RoBERTa model with traditional ML models on the imbalanced dataset.}\label{tbl8}
\begin{tabular*}{\tblwidth}{@{} p{2cm}p{1cm}p{1.5cm}p{1cm}p{1.5cm}@{} }
\toprule
Model & Training Loss  & Training Accuracy & Testing Loss & Testing Accuracy\\
\midrule
RF & 0.0834 & 98.91 & 0.2241 & 95.65\\
\midrule
XGB & 0.0684 & 98.02 & 0.1071 & 95.84\\
\midrule
SVM & 0.0531 & 98.77 & 0.0864 & 97.61\\
\midrule
LR & 0.1446 & 97.89 & 0.1707 & 96.46\\
\midrule
\textbf{RoBERTa} & \textbf{0.0281} & \textbf{98.57} & \textbf{0.0560} & \textbf{97.96}\\
\bottomrule
\end{tabular*}
\end{table}

\begin{table}[width=\linewidth,cols=5,pos=h]
\caption{Comparative analysis of the fine-tuned RoBERTa model with traditional ML models on the balanced dataset.}\label{tbl9}
\begin{tabular*}{\tblwidth}{@{} p{2cm}p{1cm}p{1.5cm}p{1cm}p{1.5cm}@{} }
\toprule
Model & Training Loss  & Training Accuracy & Testing Loss & Testing Accuracy\\
\midrule
RF & 0.0726 & 99.17 & 0.1837 & 97.09\\
\midrule
XGB & 0.0666 & 98.32 & 0.0966 & 96.49\\
\midrule
SVM & 0.0422 & 99.01 & 0.0710 & 97.90\\
\midrule
LR & 0.1343 & 98.36 & 0.1489 & 97.24\\
\midrule
\textbf{RoBERTa} & \textbf{0.0257} & \textbf{98.94} & \textbf{0.0511} & \textbf{98.45}\\
\bottomrule
\end{tabular*}
\end{table}

Finally, we have compared our Phishing Email Detection study with some previous studies in the same field in order to assess its quality. This comparison study is summarized in Table~\ref{tbl10}, which evaluates our work from several angles. These include the models used for machine and deep learning, the datasets used, and whether or not the study offers human interpretation and decision explainability for the model.

\begin{table*}[htbp]
    \caption{Explainability of Different Studies on Phishing Email Detection}\label{tbl10}
    \begin{tabularx}{\textwidth}{@{}lp{4cm}l>{\centering\arraybackslash}X@{}}
        \toprule
        Reference & Datasets & Models & Explainability Analysis \\
        \midrule
        2023\citep{jamal2023improved} & Email Spam Detection Dataset, PhishingEmailData & DistilBERT, RoBERTA & No \\
        2021\citep{alhogail2021applying} & Fraud Dataset & GCN & No \\
        2019\citep{fang2019phishing} & IWSPA-AP & THEMIS & No \\
        2023\citep{thapa2023evaluation} & IWSPA-AP, Nazario, Enron, CSIRO, Phisbowl & THEMIS, BERT & No \\
        2023\citep{atawneh2023phishing} & Phishing Email Dataset, Benign Email Dataset & BERT-LSTM & No \\
        \textbf{This Study} & \textbf{Phishing Email} & \textbf{Fine-tuned RoBERTa} & \textbf{Yes (LIME, Transformers Interpret, Hybrid LITA)} \\
        \bottomrule
    \end{tabularx}
\end{table*}

The comparison showed that our research work provides transparency in the phishing email detection task. However, some of the listed studies showed higher accuracy than our proposed model. This happened because they used different datasets in their experiment. In this study, we only focused on the phishing email dataset, whereas other existing works also included the spam datasets. Furthermore, to the best of our knowledge, none of the papers listed in Table ~\ref{tbl10} offer a thorough human explanation of the classifier's performance or clarify how the model functions in phishing email detection.

\section{Explainability Analysis}

LIME and Transformers Interpret both demonstrated how the fine-tuned RoBERTa model arrives at its prediction for phishing email detection by approximating the model’s behavior on every word of the email text. These explainability methods provided accurate results by answering “Why” the model classifies a selected text as a legitimate or phishing email. Based on the explainability, these methods assign positive and negative importance scores to each word, indicating their contribution to the predicted class. These scores highlight which word was more specifically reasonable for the predicted class. By using those scores, the explainability models visualize the whole text by highlighting the words that scored positive or negative on the prediction.

\begin{figure*}[htbp]
    \begin{subfigure}{\textwidth}
        \centering
        \includegraphics[width=0.85\textwidth]{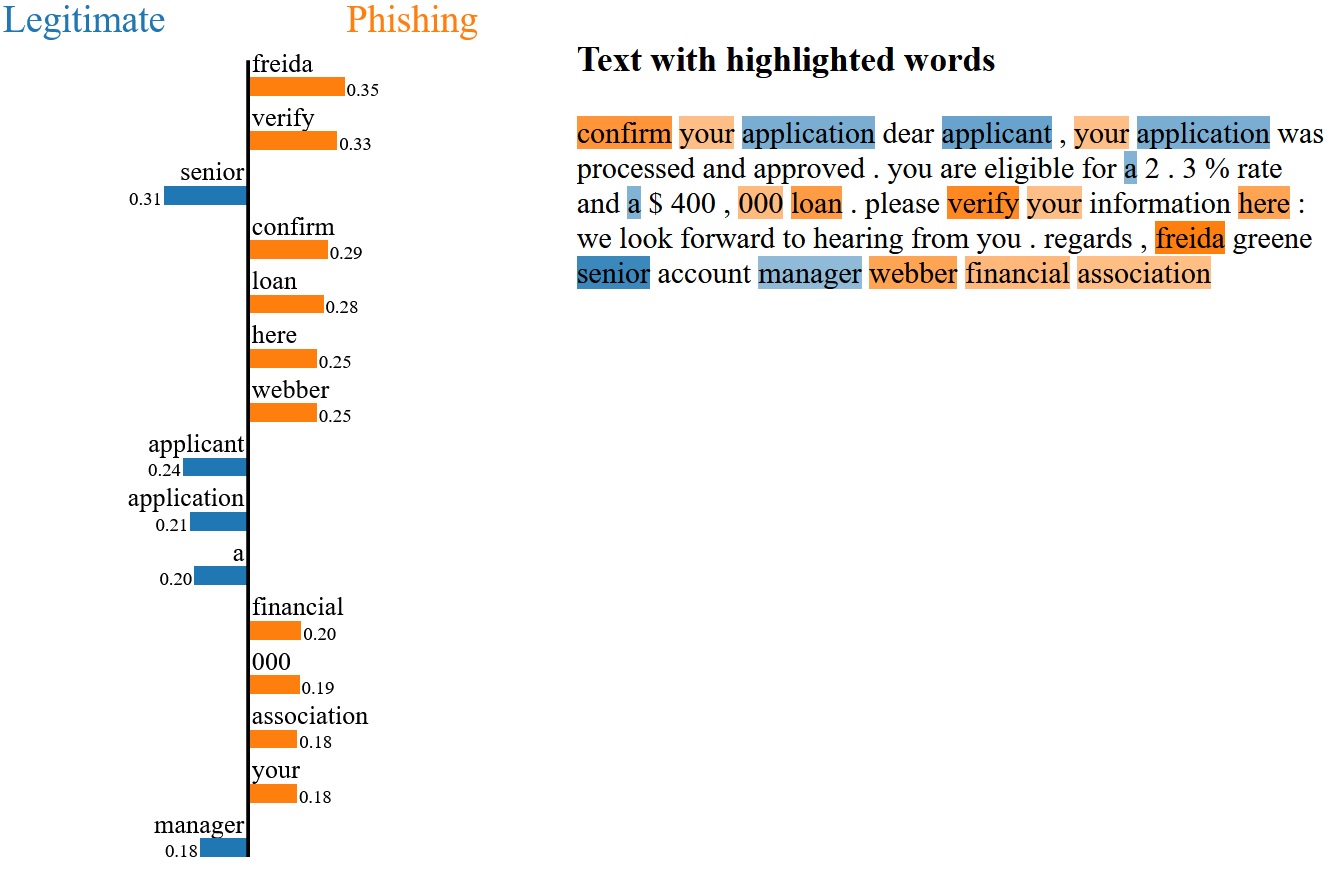}
        \caption{Explanation for the phishing sample 1}
    \end{subfigure}

    \begin{subfigure}{\textwidth}
        \centering
        \includegraphics[width=0.85\textwidth]{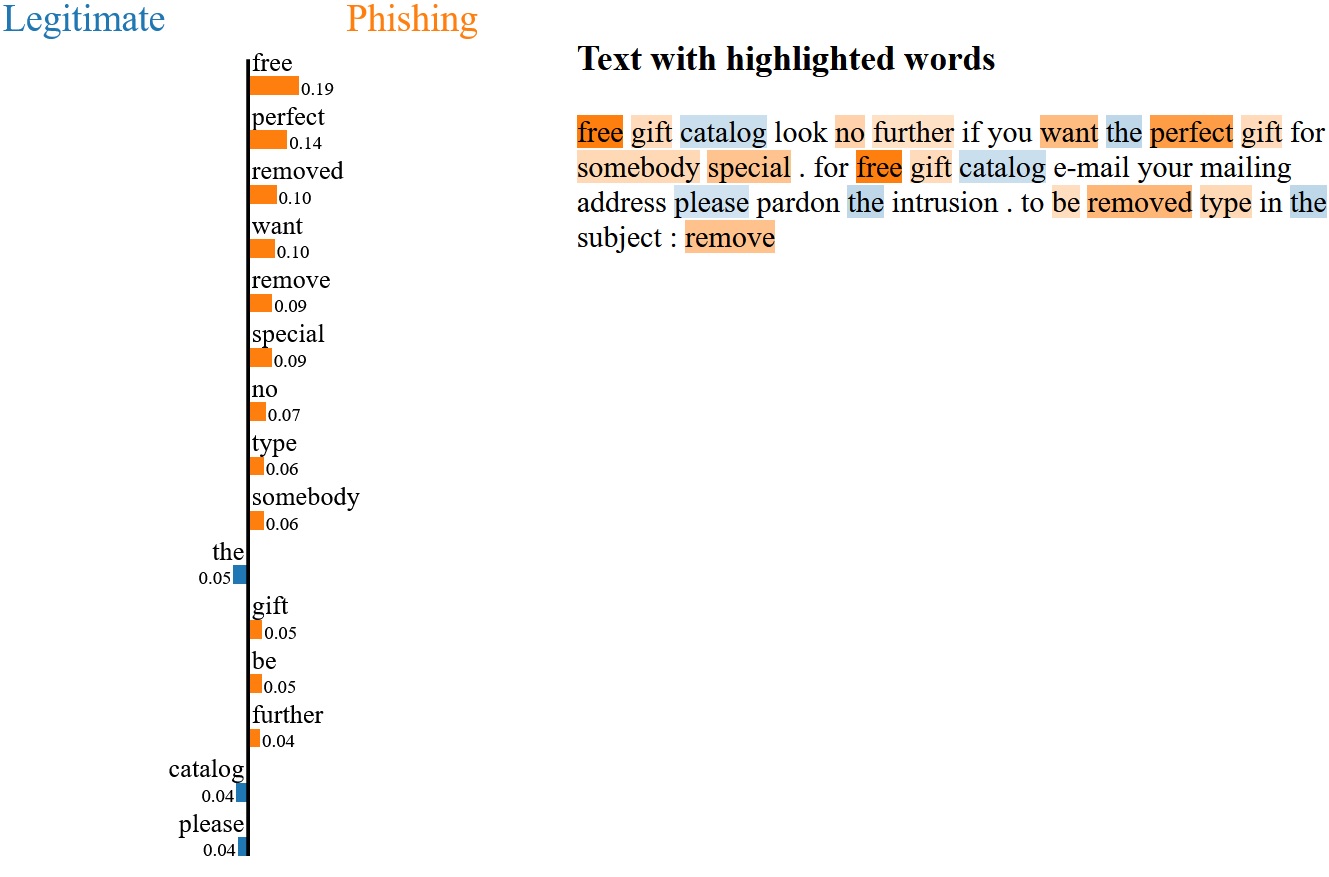}
        \caption{Explanation for the phishing sample 2}
    \end{subfigure}

    \caption{LIME Explanation for two Phishing Email Samples}
    \label{fig:LIME_Explanation}
\end{figure*}

Fig.~\ref{fig:LIME_Explanation} shows the LIME-based explanation for two phishing email samples, which the model also classified as phishing. The contributions can be seen in context by highlighting keywords in the text instances with the appropriate color. To better understand the explanation produced by the LIME explainer, presented bar plots of those phishing email samples by indicating the top 15 words ranked by their contribution to the classification. The bar length with weight indicates the strength of each word. Furthermore, to make the explanation relevant to the most likely result, we concentrated on the top-predicted label. This is especially helpful when the main goal is to comprehend the choice for the most likely class.

\begin{figure*}[htbp]
    \centering
    \begin{subfigure}{\textwidth}
        \centering
        \includegraphics[width=\textwidth]{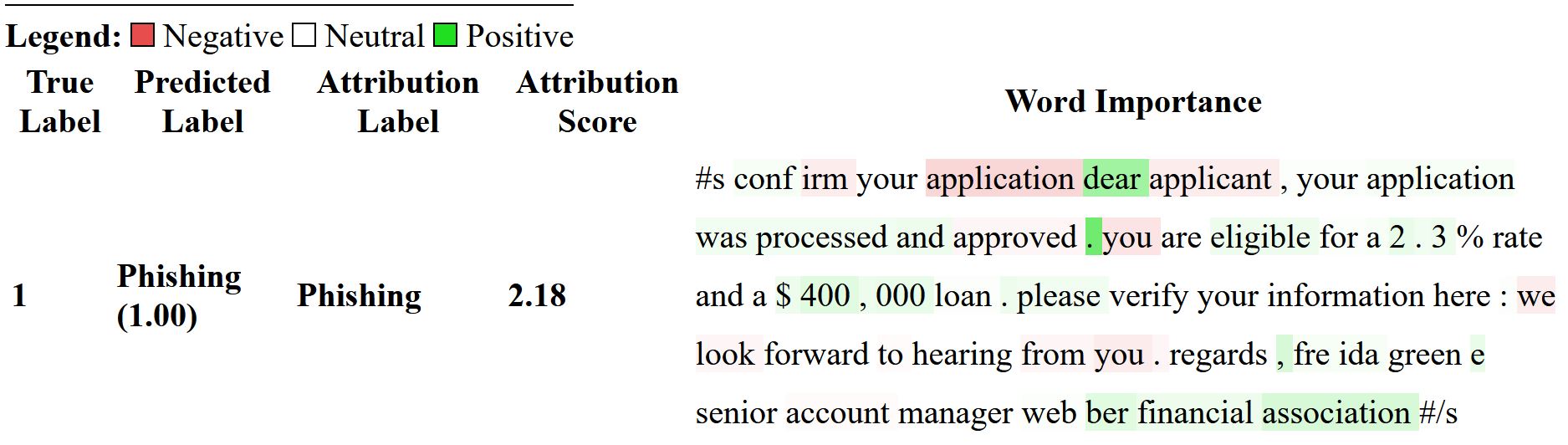}
        \caption{Explanation for sample 1}
    \end{subfigure}

    \begin{subfigure}{\textwidth}
        \centering
        \includegraphics[width=\textwidth]{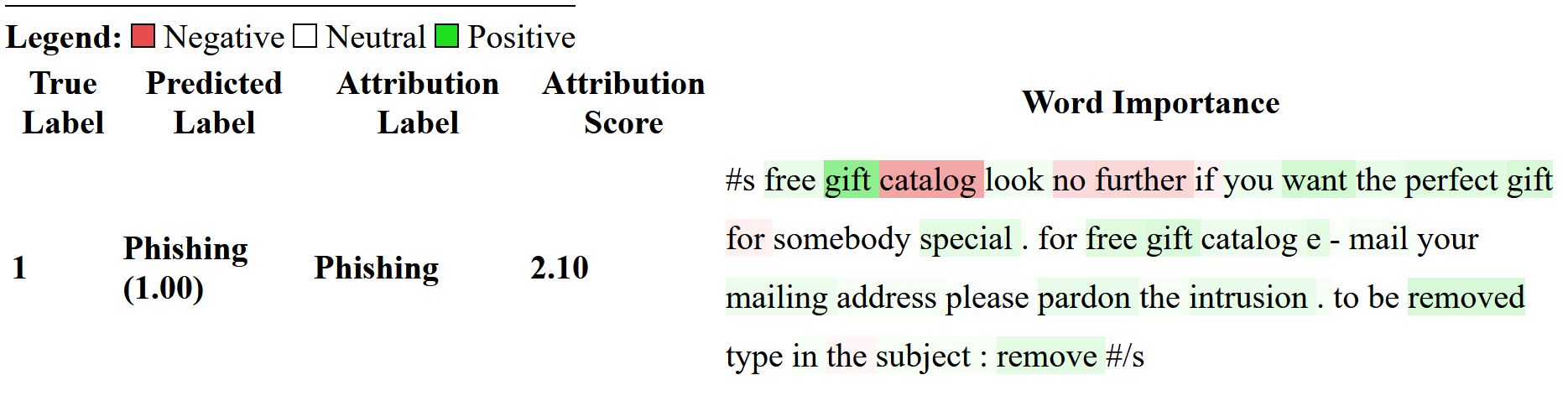}
        \caption{Explanation for sample 2}
    \end{subfigure}
    
    \caption{Transformers Interpret Explanation for two Phishing Email Samples}\label{fig:Interpret_Explanation}
\end{figure*}

In addition to LIME, we used Transformers Interpret to analyze model interpretability through comparable phishing examples. This tool examines how each token influences the model's final prediction to generate token-level attribution scores. Since RoBERTa uses a BPE tokenizer, complex or rare words are broken into smaller word units, making it easier for the model to handle out-of-vocabulary terms. Figure~\ref{fig:Interpret_Explanation} illustrates a visual interpretation for phishing email samples, where tokens highlighted in green contributed positively to the classification as phishing email, and tokens in red contributed negatively. 


According to the analysis, both explainable approaches explain how the suggested model predicts "Phishing Email" for each of the tested input samples. However, those techniques differ fundamentally in how the explanations are generated. LIME primarily uses small textual perturbations to evaluate the impact of individual words on the model's prediction across the entire text. At the same time, Transformers Interpret employs a gradient-based explanation technique to analyze the contribution of each token or sub-word to the model’s final prediction. Furthermore, in the LIME and Transformers Interpret explanation processes, the resulting quantity of tokens, words, or sub-words differs for the two phishing email samples. In both techniques, the positively contributing words are mainly responsible for the predicted class. More specifically, the presence of those tokens significantly increases the likelihood of the model predicting the sample as phishing. 

\begin{figure*}[htbp]
    \begin{subfigure}{\textwidth}
        \centering
        \includegraphics[width=0.8\textwidth]{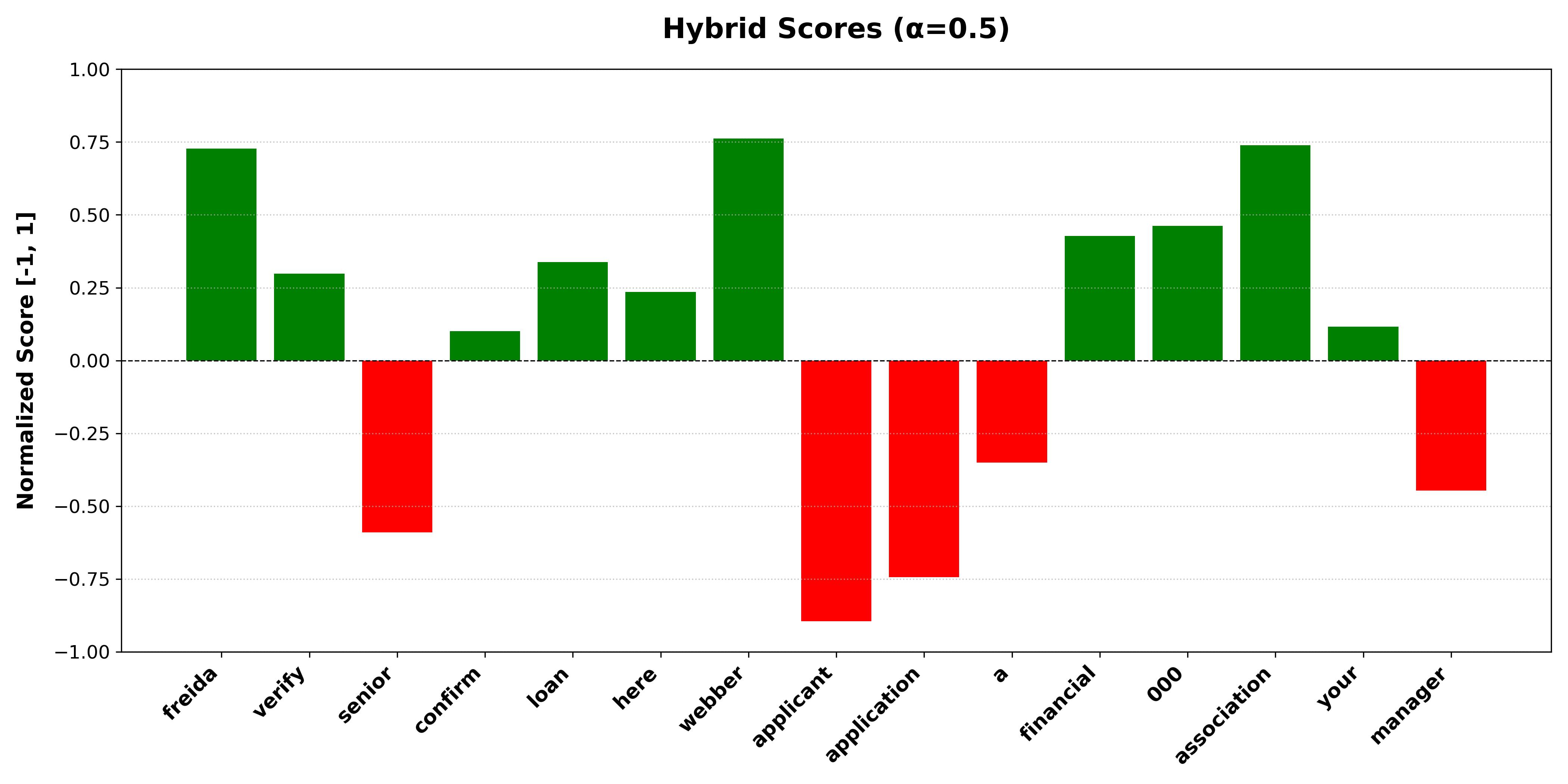}
        \caption{Explanation for the phishing sample 1}
    \end{subfigure}

    \begin{subfigure}{\textwidth}
        \centering
        \includegraphics[width=0.8\textwidth]{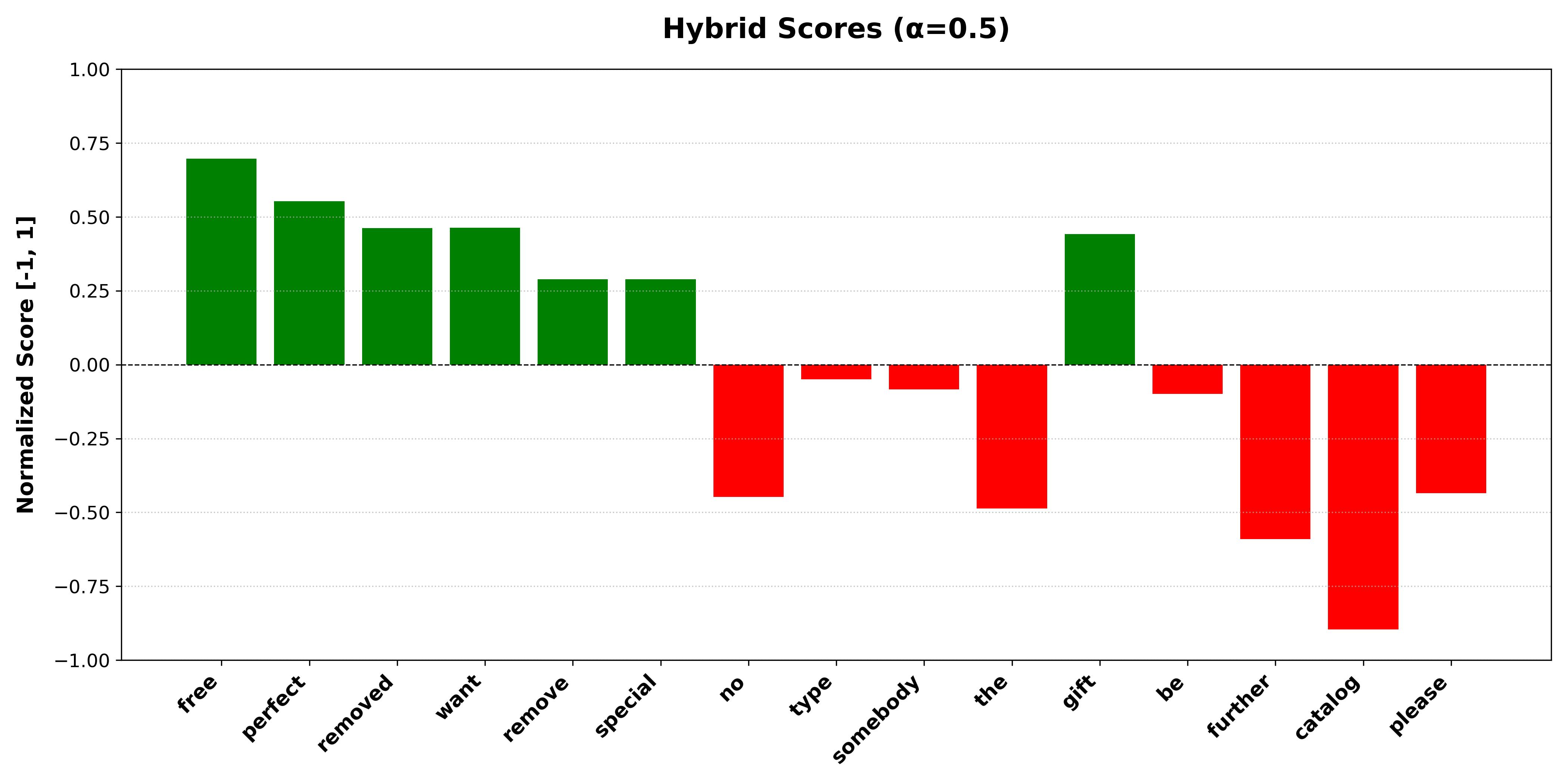}
        \caption{Explanation for the phishing sample 2}
    \end{subfigure}

    \caption{LITA Explanation for two Phishing Email Samples}
    \label{fig:LITA_Explanation}
\end{figure*}

In our study, we found that both methods can have opposite polarities for the same words. For example, one explainable method finds a word to be positive and the other negative, which can lead to confusion in the mind of end-users. To address this issue, we developed our hybrid approach, LITA, for our analysis. By integrating the complementary strengths of the two methods, we achieve a more balanced explanation. The LITA also generated explanations for two phishing samples shown in Fig.~\ref{fig:LITA_Explanation}; and Table~\ref{tblLITA1} and  Table~\ref{tblLITA2} present the top 15 words ranked by their LIME scores for each sample, along with the corresponding scores from Transformers Interpret and the hybrid LITA approach. These tables present quantitative evidence that highlights interpretability differences between both individual methods and illuminates how LITA reconciles aligned and conflicting attributions to improve clarity and trust.

\begin{table*}[htbp]
    \caption{Word-Level Explainability Analysis for Phishing Sample 1 Using LIME, Transformers Interpret, and Hybrid Scoring}\label{tblLITA1}
    \begin{tabularx}{\textwidth}{@{}lXXXXX@{}}
        \toprule
        Word & LIME Score & LIME (Norm) Score & Transformer Score & Transformer (Norm) Score & Hybrid Score \\
        \midrule
        freida        & 0.35498  & 1.00000   & 0.10828  & 0.45581   & 0.72791 \\
        verify        & 0.32802  & 0.91895   & -0.01591 & -0.32235  & 0.29830 \\
        senior        & -0.31029 & -1.00000  & 0.00691  & -0.17936  & -0.58968 \\
        confirm       & 0.29308  & 0.81391   & -0.06232 & -0.61315  & 0.10038 \\
        loan          & 0.27749  & 0.76704   & 0.02076  & -0.09258  & 0.33723 \\
        here          & 0.25293  & 0.69321   & 0.00004  & -0.22241  & 0.23540 \\
        webber        & 0.25283  & 0.69291   & 0.16803  & 0.83020   & 0.76155 \\
        applicant     & -0.24042 & -0.78995  & -0.12406 & -1.00000  & -0.89497 \\
        application   & -0.21336 & -0.70860  & -0.08878 & -0.77894  & -0.74377 \\
        a             & -0.19875 & -0.66468  & 0.02981  & -0.03587  & -0.35027 \\
        financial     & 0.19619  & 0.52263   & 0.08843  & 0.33143   & 0.42703 \\
        000           & 0.18949  & 0.50249   & 0.10274  & 0.42110   & 0.46179 \\
        association   & 0.18149  & 0.47844   & 0.19513  & 1.00000   & 0.73922 \\
        your          & 0.17999  & 0.47393   & -0.00322 & -0.24283  & 0.11555 \\
        manager       & -0.17817 & -0.60281  & -0.01086 & -0.29070  & -0.44676 \\
        \bottomrule
    \end{tabularx}
\end{table*}

\begin{table*}[htbp]
    \caption{Word-Level Explainability Analysis for Phishing Sample 2 Using LIME, Transformers Interpret, and Hybrid Scoring}\label{tblLITA2}
    \begin{tabularx}{\textwidth}{@{}lXXXXX@{}}
        \toprule
        Word & LIME Score & LIME (Norm) Score & Transformer Score & Transformer (Norm) Score & Hybrid Score \\
        \midrule
        free            & 0.18759  & 1.00000   & 0.13196  & 0.39400   & 0.69700 \\
        perfect         & 0.14421  & 0.64174   & 0.15117  & 0.46549   & 0.55361 \\
        removed         & 0.10480  & 0.31627   & 0.18932  & 0.60745   & 0.46186 \\
        want            & 0.09803  & 0.26035   & 0.20522  & 0.66662   & 0.46349 \\
        remove          & 0.08880  & 0.18413   & 0.13166  & 0.39289   & 0.28851 \\
        special         & 0.08735  & 0.17215   & 0.13495  & 0.40513   & 0.28864 \\
        no              & 0.06542  & -0.00896  & -0.21209 & -0.88628  & -0.44762 \\
        type            & 0.05757  & -0.07379  & 0.01917  & -0.02571  & -0.04975 \\
        somebody        & 0.05757  & -0.07379  & 0.00061  & -0.09478  & -0.08429 \\
        the             & -0.05458 & -1.00000  & 0.03316  & 0.02635   & -0.48683 \\
        gift            & 0.05245  & -0.11608  & 0.29481  & 1.00000   & 0.44196 \\
        be              & 0.04975  & -0.13837  & 0.00989  & -0.06025  & -0.09931 \\
        further         & 0.04455  & -0.18132  & -0.24265 & -1.00000  & -0.59066 \\
        catalog         & -0.04401 & -0.91271  & -0.21059 & -0.88070  & -0.89670 \\
        please          & -0.03814 & -0.86423  & 0.02428  & -0.00670  & -0.43546 \\
        \bottomrule
    \end{tabularx}
\end{table*}

\section{Discussion}

The results of both an imbalanced and a balanced dataset depict an enhancement in the performance of the proposed transformer-based fine-tuned RoBERTa model. The fine-tuned version of the RoBERTa model showed 98.57\% training and 97.96\% testing accuracy on an imbalanced dataset, whereas this model showed 98.94\% training and 98.45\% testing accuracy on the balanced dataset. Although these results show that the model's performance on a balanced dataset is quite better than on an imbalanced dataset, the difference is not so high, approximately 0.49\%. This demonstrated that the fine-tuned version performs well in an imbalanced dataset. Additionally, the performance shows that the optimized model performed better than other previous studies.

Additionally, this research work also addressed the questions “How was the LLM model performed?" and “Why was it performed?" using XAI, as motivated by the recently published paper by Sarker \citep{sarker2024llm}. The author highlighted the potentiality of the LLMs by addressing the technical, ethical, and societal challenges, focusing on fairness, transparency, transparency, explainability, and trust. By summarizing and understanding this potential research scope with direction, two well-known explainable and interpretable techniques are utilized in our research to understand the model’s decision. It is more difficult to understand how RoBERTa makes its predictions because, as we know, it is a complex model. These AI techniques aided in gaining an understanding of the model's decision-making process. Both LIME and Transformers Interpret highlighted the areas of the text that are more responsible for the classification of phishing or legitimate emails. This highlights areas worked with words or sub-words, making the decision more understandable for humans, and users can also comprehend it easily. Mainly, those techniques calculated each word or sub-words numeric value or weight, where some words indicate positive value and some negative. Words with calculated higher positive values are more important for classifying the predicted class, while those with more negative values do not support the predicted class. This experimental analysis showed the transparency of the model in this phishing email classification task. Although those methods show almost similar explanations, but in some word cases their explanation not agreed due to their different functionality. But this can create serious confusion which one solved in this work, with LITA.

\section{Conclusion}
This study explores the detection of phishing emails using a transformer-based language model. It also demonstrates the effectiveness of employing state-of-the-art natural language processing methods to improve the accuracy and efficiency of identifying phishing emails. The overall experiment was conducted with both balanced and imbalanced datasets. We achieved promising results in distinguishing between phishing and legitimate emails by utilizing RoBERTa's contextual language understanding capabilities. To accomplish this, the model was optimized by adjusting key parameters such as batch size, optimization function, learning rate, and through fine-tuning. This fine-tuned version of RoBERTa obtained a 98.45\% testing accuracy in the phishing email classification task. Additionally, to improve interpretability and foster greater trust in the model’s explanations, we introduced LITA, a hybrid explanation approach that combines LIME and Transformers Interpret. This makes the decision-making process more transparent and understandable for users, promoting comprehension and trust. Overall, the model’s explainability reflects the robustness of the proposed system.

The field of LLMs has experienced quick and dynamic development due to the substantial investment made by both businesses and consumers. New open-source models are being produced almost every day, demonstrating the ongoing development and innovation in this field. In future work, we plan to explore various LLMs, including LLaMA, Gemini, Falcon, Mistral, and others, to assess their effectiveness in phishing email detection. We will extend our research to multilingual datasets to improve our model’s performance across different languages and cultural contexts. To further enhance the robustness and accuracy of our algorithms, we also plan to create a new dataset comprising phishing emails. Additionally, we aim to integrate our models into real-world cybersecurity systems, partnering with industry players to evaluate their real-time effectiveness and refine them based on practical feedback. Furthermore, we will conduct more in-depth analysis of the models' decision-making processes to ensure robustness by advancing explainable AI techniques.

\printcredits

\section*{Data Availability Statement}

The research dataset which one is used in this study is available at Kaggle: ~\url{https://www.kaggle.com/datasets/subhajournal/phishingemails}.


\begin{thebibliography}{50}
\expandafter\ifx\csname natexlab\endcsname\relax\def\natexlab#1{#1}\fi
\providecommand{\url}[1]{\texttt{#1}}
\providecommand{\href}[2]{#2}
\providecommand{\path}[1]{#1}
\providecommand{\DOIprefix}{doi:}
\providecommand{\ArXivprefix}{arXiv:}
\providecommand{\URLprefix}{URL: }
\providecommand{\Pubmedprefix}{pmid:}
\providecommand{\doi}[1]{\href{http://dx.doi.org/#1}{\path{#1}}}
\providecommand{\Pubmed}[1]{\href{pmid:#1}{\path{#1}}}
\providecommand{\bibinfo}[2]{#2}
\ifx\xfnm\relax \def\xfnm[#1]{\unskip,\space#1}\fi
\bibitem[{Salloum et~al.(2021)Salloum, Gaber, Vadera, and Shaalan}]{salloum2021phishing}
\bibinfo{author}{S.~Salloum}, \bibinfo{author}{T.~Gaber}, \bibinfo{author}{S.~Vadera}, \bibinfo{author}{K.~Shaalan},
\newblock \bibinfo{title}{Phishing email detection using natural language processing techniques: a literature survey},
\newblock \bibinfo{journal}{Procedia Computer Science} \bibinfo{volume}{189} (\bibinfo{year}{2021}) \bibinfo{pages}{19--28}.
\bibitem[{Almomani et~al.(2013)Almomani, Gupta, Atawneh, Meulenberg, and Almomani}]{almomani2013survey}
\bibinfo{author}{A.~Almomani}, \bibinfo{author}{B.~B. Gupta}, \bibinfo{author}{S.~Atawneh}, \bibinfo{author}{A.~Meulenberg}, \bibinfo{author}{E.~Almomani},
\newblock \bibinfo{title}{A survey of phishing email filtering techniques},
\newblock \bibinfo{journal}{IEEE communications surveys \& tutorials} \bibinfo{volume}{15} (\bibinfo{year}{2013}) \bibinfo{pages}{2070--2090}.
\bibitem[{Basit et~al.(2021)Basit, Zafar, Liu, Javed, Jalil, and Kifayat}]{basit2021comprehensive}
\bibinfo{author}{A.~Basit}, \bibinfo{author}{M.~Zafar}, \bibinfo{author}{X.~Liu}, \bibinfo{author}{A.~R. Javed}, \bibinfo{author}{Z.~Jalil}, \bibinfo{author}{K.~Kifayat},
\newblock \bibinfo{title}{A comprehensive survey of ai-enabled phishing attacks detection techniques},
\newblock \bibinfo{journal}{Telecommunication Systems} \bibinfo{volume}{76} (\bibinfo{year}{2021}) \bibinfo{pages}{139--154}.
\bibitem[{Sarker(2023)}]{sarker2023machine}
\bibinfo{author}{I.~H. Sarker},
\newblock \bibinfo{title}{Machine learning for intelligent data analysis and automation in cybersecurity: current and future prospects},
\newblock \bibinfo{journal}{Annals of Data Science} \bibinfo{volume}{10} (\bibinfo{year}{2023}) \bibinfo{pages}{1473--1498}.
\bibitem[{Han et~al.(2021)Han, Xiao, Wu, Guo, Xu, and Wang}]{han2021transformer}
\bibinfo{author}{K.~Han}, \bibinfo{author}{A.~Xiao}, \bibinfo{author}{E.~Wu}, \bibinfo{author}{J.~Guo}, \bibinfo{author}{C.~Xu}, \bibinfo{author}{Y.~Wang},
\newblock \bibinfo{title}{Transformer in transformer},
\newblock \bibinfo{journal}{Advances in Neural Information Processing Systems} \bibinfo{volume}{34} (\bibinfo{year}{2021}) \bibinfo{pages}{15908--15919}.
\bibitem[{Jamal et~al.(2023)Jamal, Wimmer, and Sarker}]{jamal2023improved}
\bibinfo{author}{S.~Jamal}, \bibinfo{author}{H.~Wimmer}, \bibinfo{author}{I.~Sarker},
\newblock \bibinfo{title}{An improved transformer-based model for detecting phishing, spam, and ham: A large language model approach},
\newblock \bibinfo{journal}{arXiv preprint arXiv:2311.04913}  (\bibinfo{year}{2023}).
\bibitem[{Zhao et~al.(2023)Zhao, Zhou, Li, Tang, Wang, Hou, Min, Zhang, Zhang, Dong et~al.}]{zhao2023survey}
\bibinfo{author}{W.~X. Zhao}, \bibinfo{author}{K.~Zhou}, \bibinfo{author}{J.~Li}, \bibinfo{author}{T.~Tang}, \bibinfo{author}{X.~Wang}, \bibinfo{author}{Y.~Hou}, \bibinfo{author}{Y.~Min}, \bibinfo{author}{B.~Zhang}, \bibinfo{author}{J.~Zhang}, \bibinfo{author}{Z.~Dong}, et~al.,
\newblock \bibinfo{title}{A survey of large language models},
\newblock \bibinfo{journal}{arXiv preprint arXiv:2303.18223}  (\bibinfo{year}{2023}).
\bibitem[{Yao et~al.(2023)Yao, Duan, Xu, Cai, Sun, and Zhang}]{yao2023survey}
\bibinfo{author}{Y.~Yao}, \bibinfo{author}{J.~Duan}, \bibinfo{author}{K.~Xu}, \bibinfo{author}{Y.~Cai}, \bibinfo{author}{E.~Sun}, \bibinfo{author}{Y.~Zhang},
\newblock \bibinfo{title}{A survey on large language model (llm) security and privacy: The good, the bad, and the ugly},
\newblock \bibinfo{journal}{arXiv preprint arXiv:2312.02003}  (\bibinfo{year}{2023}).
\bibitem[{Koroteev(2021)}]{koroteev2021bert}
\bibinfo{author}{M.~Koroteev},
\newblock \bibinfo{title}{Bert: a review of applications in natural language processing and understanding},
\newblock \bibinfo{journal}{arXiv preprint arXiv:2103.11943}  (\bibinfo{year}{2021}).
\bibitem[{Singh et~al.(2021)Singh, Jakhar, and Pandey}]{singh2021sentiment}
\bibinfo{author}{M.~Singh}, \bibinfo{author}{A.~K. Jakhar}, \bibinfo{author}{S.~Pandey},
\newblock \bibinfo{title}{Sentiment analysis on the impact of coronavirus in social life using the bert model},
\newblock \bibinfo{journal}{Social Network Analysis and Mining} \bibinfo{volume}{11} (\bibinfo{year}{2021}) \bibinfo{pages}{33}.
\bibitem[{Devlin et~al.(2018)Devlin, Chang, Lee, and Toutanova}]{devlin2018bert}
\bibinfo{author}{J.~Devlin}, \bibinfo{author}{M.-W. Chang}, \bibinfo{author}{K.~Lee}, \bibinfo{author}{K.~Toutanova},
\newblock \bibinfo{title}{Bert: Pre-training of deep bidirectional transformers for language understanding},
\newblock \bibinfo{journal}{arXiv preprint arXiv:1810.04805}  (\bibinfo{year}{2018}).
\bibitem[{Khadhraoui et~al.(2022)Khadhraoui, Bellaaj, Ammar, Hamam, and Jmaiel}]{khadhraoui2022survey}
\bibinfo{author}{M.~Khadhraoui}, \bibinfo{author}{H.~Bellaaj}, \bibinfo{author}{M.~B. Ammar}, \bibinfo{author}{H.~Hamam}, \bibinfo{author}{M.~Jmaiel},
\newblock \bibinfo{title}{Survey of bert-base models for scientific text classification: Covid-19 case study},
\newblock \bibinfo{journal}{Applied Sciences} \bibinfo{volume}{12} (\bibinfo{year}{2022}) \bibinfo{pages}{2891}.
\bibitem[{Lan et~al.(2019)Lan, Chen, Goodman, Gimpel, Sharma, and Soricut}]{lan2019albert}
\bibinfo{author}{Z.~Lan}, \bibinfo{author}{M.~Chen}, \bibinfo{author}{S.~Goodman}, \bibinfo{author}{K.~Gimpel}, \bibinfo{author}{P.~Sharma}, \bibinfo{author}{R.~Soricut},
\newblock \bibinfo{title}{Albert: A lite bert for self-supervised learning of language representations},
\newblock \bibinfo{journal}{arXiv preprint arXiv:1909.11942}  (\bibinfo{year}{2019}).
\bibitem[{Liu et~al.(2019)Liu, Ott, Goyal, Du, Joshi, Chen, Levy, Lewis, Zettlemoyer, and Stoyanov}]{liu2019roberta}
\bibinfo{author}{Y.~Liu}, \bibinfo{author}{M.~Ott}, \bibinfo{author}{N.~Goyal}, \bibinfo{author}{J.~Du}, \bibinfo{author}{M.~Joshi}, \bibinfo{author}{D.~Chen}, \bibinfo{author}{O.~Levy}, \bibinfo{author}{M.~Lewis}, \bibinfo{author}{L.~Zettlemoyer}, \bibinfo{author}{V.~Stoyanov},
\newblock \bibinfo{title}{Roberta: A robustly optimized bert pretraining approach},
\newblock \bibinfo{journal}{arXiv preprint arXiv:1907.11692}  (\bibinfo{year}{2019}).
\bibitem[{Sanh et~al.(2019)Sanh, Debut, Chaumond, and Wolf}]{sanh2019distilbert}
\bibinfo{author}{V.~Sanh}, \bibinfo{author}{L.~Debut}, \bibinfo{author}{J.~Chaumond}, \bibinfo{author}{T.~Wolf},
\newblock \bibinfo{title}{Distilbert, a distilled version of bert: smaller, faster, cheaper and lighter},
\newblock \bibinfo{journal}{arXiv preprint arXiv:1910.01108}  (\bibinfo{year}{2019}).
\bibitem[{Xu et~al.(2019)Xu, Uszkoreit, Du, Fan, Zhao, and Zhu}]{xu2019explainable}
\bibinfo{author}{F.~Xu}, \bibinfo{author}{H.~Uszkoreit}, \bibinfo{author}{Y.~Du}, \bibinfo{author}{W.~Fan}, \bibinfo{author}{D.~Zhao}, \bibinfo{author}{J.~Zhu},
\newblock \bibinfo{title}{Explainable ai: A brief survey on history, research areas, approaches and challenges},
\newblock in: \bibinfo{booktitle}{Natural Language Processing and Chinese Computing: 8th CCF International Conference, NLPCC 2019, Dunhuang, China, October 9--14, 2019, Proceedings, Part II 8}, \bibinfo{organization}{Springer}, \bibinfo{year}{2019}, pp. \bibinfo{pages}{563--574}.
\bibitem[{Hoffman et~al.(2018)Hoffman, Mueller, Klein, and Litman}]{hoffman2018metrics}
\bibinfo{author}{R.~R. Hoffman}, \bibinfo{author}{S.~T. Mueller}, \bibinfo{author}{G.~Klein}, \bibinfo{author}{J.~Litman},
\newblock \bibinfo{title}{Metrics for explainable ai: Challenges and prospects},
\newblock \bibinfo{journal}{arXiv preprint arXiv:1812.04608}  (\bibinfo{year}{2018}).
\bibitem[{Anan et~al.(2023)Anan, Apon, Hossain, Modhu, Mondal, and Alam}]{anan2023interpretable}
\bibinfo{author}{R.~Anan}, \bibinfo{author}{T.~S. Apon}, \bibinfo{author}{Z.~T. Hossain}, \bibinfo{author}{E.~A. Modhu}, \bibinfo{author}{S.~Mondal}, \bibinfo{author}{M.~G.~R. Alam},
\newblock \bibinfo{title}{Interpretable bangla sarcasm detection using bert and explainable ai},
\newblock in: \bibinfo{booktitle}{2023 IEEE 13th Annual Computing and Communication Workshop and Conference (CCWC)}, \bibinfo{organization}{IEEE}, \bibinfo{year}{2023}, pp. \bibinfo{pages}{1272--1278}.
\bibitem[{Sarker(2024)}]{sarker2024ai}
\bibinfo{author}{I.~H. Sarker}, \bibinfo{title}{AI-driven cybersecurity and threat intelligence: cyber automation, intelligent decision-making and explainability}, \bibinfo{publisher}{Springer}, \bibinfo{year}{2024}.
\bibitem[{Sarker et~al.(2024)Sarker, Janicke, Mohsin, Gill, and Maglaras}]{sarker2024explainable}
\bibinfo{author}{I.~H. Sarker}, \bibinfo{author}{H.~Janicke}, \bibinfo{author}{A.~Mohsin}, \bibinfo{author}{A.~Gill}, \bibinfo{author}{L.~Maglaras},
\newblock \bibinfo{title}{Explainable ai for cybersecurity automation, intelligence and trustworthiness in digital twin: Methods, taxonomy, challenges and prospects},
\newblock \bibinfo{journal}{ICT Express}  (\bibinfo{year}{2024}).
\bibitem[{Uddin et~al.(2025)Uddin, Islam, Maglaras, Janicke, and Sarker}]{uddin2025explainabledetector}
\bibinfo{author}{M.~A. Uddin}, \bibinfo{author}{M.~N. Islam}, \bibinfo{author}{L.~Maglaras}, \bibinfo{author}{H.~Janicke}, \bibinfo{author}{I.~H. Sarker},
\newblock \bibinfo{title}{Explainabledetector: Exploring transformer-based language modeling approach for sms spam detection with explainability analysis},
\newblock \bibinfo{journal}{Digital Communications and Networks}  (\bibinfo{year}{2025}).
\bibitem[{Apruzzese et~al.(2023)Apruzzese, Laskov, Montes~de Oca, Mallouli, Brdalo~Rapa, Grammatopoulos, and Di~Franco}]{apruzzese2023role}
\bibinfo{author}{G.~Apruzzese}, \bibinfo{author}{P.~Laskov}, \bibinfo{author}{E.~Montes~de Oca}, \bibinfo{author}{W.~Mallouli}, \bibinfo{author}{L.~Brdalo~Rapa}, \bibinfo{author}{A.~V. Grammatopoulos}, \bibinfo{author}{F.~Di~Franco},
\newblock \bibinfo{title}{The role of machine learning in cybersecurity},
\newblock \bibinfo{journal}{Digital Threats: Research and Practice} \bibinfo{volume}{4} (\bibinfo{year}{2023}) \bibinfo{pages}{1--38}.
\bibitem[{Yasin and Abuhasan(2016)}]{yasin2016intelligent}
\bibinfo{author}{A.~Yasin}, \bibinfo{author}{A.~Abuhasan},
\newblock \bibinfo{title}{An intelligent classification model for phishing email detection},
\newblock \bibinfo{journal}{arXiv preprint arXiv:1608.02196}  (\bibinfo{year}{2016}).
\bibitem[{Harikrishnan et~al.(2018)Harikrishnan, Vinayakumar, and Soman}]{harikrishnan2018machine}
\bibinfo{author}{N.~Harikrishnan}, \bibinfo{author}{R.~Vinayakumar}, \bibinfo{author}{K.~Soman},
\newblock \bibinfo{title}{A machine learning approach towards phishing email detection},
\newblock in: \bibinfo{booktitle}{Proceedings of the Anti-Phishing Pilot at ACM International Workshop on Security and Privacy Analytics (IWSPA AP)}, volume \bibinfo{volume}{2013}, \bibinfo{year}{2018}, pp. \bibinfo{pages}{455--468}.
\bibitem[{Hamid et~al.(2013)Hamid, Abawajy, and Kim}]{hamid2013using}
\bibinfo{author}{I.~R.~A. Hamid}, \bibinfo{author}{J.~Abawajy}, \bibinfo{author}{T.~Kim},
\newblock \bibinfo{title}{Using feature selection and classification scheme for automating phishing email detection},
\newblock \bibinfo{journal}{Studies in Informatics and Control} \bibinfo{volume}{22} (\bibinfo{year}{2013}).
\bibitem[{Zamir et~al.(2020)Zamir, Khan, Iqbal, Yousaf, Aslam, Anjum, and Hamdani}]{zamir2020phishing}
\bibinfo{author}{A.~Zamir}, \bibinfo{author}{H.~U. Khan}, \bibinfo{author}{T.~Iqbal}, \bibinfo{author}{N.~Yousaf}, \bibinfo{author}{F.~Aslam}, \bibinfo{author}{A.~Anjum}, \bibinfo{author}{M.~Hamdani},
\newblock \bibinfo{title}{Phishing web site detection using diverse machine learning algorithms},
\newblock \bibinfo{journal}{The Electronic Library} \bibinfo{volume}{38} (\bibinfo{year}{2020}) \bibinfo{pages}{65--80}.
\bibitem[{Almomani et~al.(2013)Almomani, Gupta, Wan, Altaher, and Manickam}]{almomani2013phishing}
\bibinfo{author}{A.~Almomani}, \bibinfo{author}{B.~B. Gupta}, \bibinfo{author}{T.-C. Wan}, \bibinfo{author}{A.~Altaher}, \bibinfo{author}{S.~Manickam},
\newblock \bibinfo{title}{Phishing dynamic evolving neural fuzzy framework for online detection zero-day phishing email},
\newblock \bibinfo{journal}{arXiv preprint arXiv:1302.0629}  (\bibinfo{year}{2013}).
\bibitem[{Salhi et~al.(2021)Salhi, Tari, and Kechadi}]{salhi2021email}
\bibinfo{author}{D.~E. Salhi}, \bibinfo{author}{A.~Tari}, \bibinfo{author}{M.~T. Kechadi},
\newblock \bibinfo{title}{Email classification for forensic analysis by information gain technique},
\newblock \bibinfo{journal}{International Journal of Software Science and Computational Intelligence (IJSSCI)} \bibinfo{volume}{13} (\bibinfo{year}{2021}) \bibinfo{pages}{40--53}.
\bibitem[{Alhogail and Alsabih(2021)}]{alhogail2021applying}
\bibinfo{author}{A.~Alhogail}, \bibinfo{author}{A.~Alsabih},
\newblock \bibinfo{title}{Applying machine learning and natural language processing to detect phishing email},
\newblock \bibinfo{journal}{Computers \& Security} \bibinfo{volume}{110} (\bibinfo{year}{2021}) \bibinfo{pages}{102414}.
\bibitem[{Brindha et~al.(2023)Brindha, Nandagopal, Azath, Sathana, Joshi, and Kim}]{brindha2023intelligent}
\bibinfo{author}{R.~Brindha}, \bibinfo{author}{S.~Nandagopal}, \bibinfo{author}{H.~Azath}, \bibinfo{author}{V.~Sathana}, \bibinfo{author}{G.~P. Joshi}, \bibinfo{author}{S.~W. Kim},
\newblock \bibinfo{title}{Intelligent deep learning based cybersecurity phishing email detection and classification.},
\newblock \bibinfo{journal}{Computers, Materials \& Continua} \bibinfo{volume}{74} (\bibinfo{year}{2023}).
\bibitem[{Dewis and Viana(2022)}]{dewis2022phish}
\bibinfo{author}{M.~Dewis}, \bibinfo{author}{T.~Viana},
\newblock \bibinfo{title}{Phish responder: A hybrid machine learning approach to detect phishing and spam emails},
\newblock \bibinfo{journal}{Applied System Innovation} \bibinfo{volume}{5} (\bibinfo{year}{2022}) \bibinfo{pages}{73}.
\bibitem[{Fang et~al.(2019)Fang, Zhang, Huang, Liu, and Yang}]{fang2019phishing}
\bibinfo{author}{Y.~Fang}, \bibinfo{author}{C.~Zhang}, \bibinfo{author}{C.~Huang}, \bibinfo{author}{L.~Liu}, \bibinfo{author}{Y.~Yang},
\newblock \bibinfo{title}{Phishing email detection using improved rcnn model with multilevel vectors and attention mechanism},
\newblock \bibinfo{journal}{IEEE Access} \bibinfo{volume}{7} (\bibinfo{year}{2019}) \bibinfo{pages}{56329--56340}.
\bibitem[{Zhang and Li(2017)}]{zhang2017phishing}
\bibinfo{author}{J.~Zhang}, \bibinfo{author}{X.~Li},
\newblock \bibinfo{title}{Phishing detection method based on borderline-smote deep belief network},
\newblock in: \bibinfo{booktitle}{Security, Privacy, and Anonymity in Computation, Communication, and Storage: SpaCCS 2017 International Workshops, Guangzhou, China, December 12-15, 2017, Proceedings 10}, \bibinfo{organization}{Springer}, \bibinfo{year}{2017}, pp. \bibinfo{pages}{45--53}.
\bibitem[{Bahnsen et~al.(2017)Bahnsen, Bohorquez, Villegas, Vargas, and Gonz{\'a}lez}]{bahnsen2017classifying}
\bibinfo{author}{A.~C. Bahnsen}, \bibinfo{author}{E.~C. Bohorquez}, \bibinfo{author}{S.~Villegas}, \bibinfo{author}{J.~Vargas}, \bibinfo{author}{F.~A. Gonz{\'a}lez},
\newblock \bibinfo{title}{Classifying phishing urls using recurrent neural networks},
\newblock in: \bibinfo{booktitle}{2017 APWG symposium on electronic crime research (eCrime)}, \bibinfo{organization}{IEEE}, \bibinfo{year}{2017}, pp. \bibinfo{pages}{1--8}.
\bibitem[{Smadi et~al.(2018)Smadi, Aslam, and Zhang}]{smadi2018detection}
\bibinfo{author}{S.~Smadi}, \bibinfo{author}{N.~Aslam}, \bibinfo{author}{L.~Zhang},
\newblock \bibinfo{title}{Detection of online phishing email using dynamic evolving neural network based on reinforcement learning},
\newblock \bibinfo{journal}{Decision Support Systems} \bibinfo{volume}{107} (\bibinfo{year}{2018}) \bibinfo{pages}{88--102}.
\bibitem[{Thapa et~al.(2023)Thapa, Tang, Abuadbba, Gao, Camtepe, Nepal, Almashor, and Zheng}]{thapa2023evaluation}
\bibinfo{author}{C.~Thapa}, \bibinfo{author}{J.~W. Tang}, \bibinfo{author}{A.~Abuadbba}, \bibinfo{author}{Y.~Gao}, \bibinfo{author}{S.~Camtepe}, \bibinfo{author}{S.~Nepal}, \bibinfo{author}{M.~Almashor}, \bibinfo{author}{Y.~Zheng},
\newblock \bibinfo{title}{Evaluation of federated learning in phishing email detection},
\newblock \bibinfo{journal}{Sensors} \bibinfo{volume}{23} (\bibinfo{year}{2023}) \bibinfo{pages}{4346}.
\bibitem[{Atawneh and Aljehani(2023)}]{atawneh2023phishing}
\bibinfo{author}{S.~Atawneh}, \bibinfo{author}{H.~Aljehani},
\newblock \bibinfo{title}{Phishing email detection model using deep learning},
\newblock \bibinfo{journal}{Electronics} \bibinfo{volume}{12} (\bibinfo{year}{2023}) \bibinfo{pages}{4261}.
\bibitem[{Jiao et~al.(2019)Jiao, Yin, Shang, Jiang, Chen, Li, Wang, and Liu}]{jiao2019tinybert}
\bibinfo{author}{X.~Jiao}, \bibinfo{author}{Y.~Yin}, \bibinfo{author}{L.~Shang}, \bibinfo{author}{X.~Jiang}, \bibinfo{author}{X.~Chen}, \bibinfo{author}{L.~Li}, \bibinfo{author}{F.~Wang}, \bibinfo{author}{Q.~Liu},
\newblock \bibinfo{title}{Tinybert: Distilling bert for natural language understanding},
\newblock \bibinfo{journal}{arXiv preprint arXiv:1909.10351}  (\bibinfo{year}{2019}).
\bibitem[{Lee et~al.(2020)Lee, Saxe, and Harang}]{lee2020catbert}
\bibinfo{author}{Y.~Lee}, \bibinfo{author}{J.~Saxe}, \bibinfo{author}{R.~Harang},
\newblock \bibinfo{title}{Catbert: Context-aware tiny bert for detecting social engineering emails},
\newblock \bibinfo{journal}{arXiv preprint arXiv:2010.03484}  (\bibinfo{year}{2020}).
\bibitem[{Songailait{\.e} et~al.(2023)Songailait{\.e}, Kankevi{\v{c}}i{\=u}t{\.e}, Zhyhun, and Mandravickait{\.e}}]{songailaite2023bert}
\bibinfo{author}{M.~Songailait{\.e}}, \bibinfo{author}{E.~Kankevi{\v{c}}i{\=u}t{\.e}}, \bibinfo{author}{B.~Zhyhun}, \bibinfo{author}{J.~Mandravickait{\.e}},
\newblock \bibinfo{title}{Bert-based models for phishing detection},
\newblock in: \bibinfo{booktitle}{CEUR Workshop proceedings: IVUS 2023: Proceedings of the 28th international conference on Information Society and University Studies, Kaunas, Lithuania, May 12, 2023.}, volume \bibinfo{volume}{3575}, \bibinfo{organization}{Aachen: CEUR-WS}, \bibinfo{year}{2023}, pp. \bibinfo{pages}{34--44}.
\bibitem[{Wang et~al.(2023)Wang, Zhu, Xu, Qin, Ren, and Ma}]{wang2023large}
\bibinfo{author}{Y.~Wang}, \bibinfo{author}{W.~Zhu}, \bibinfo{author}{H.~Xu}, \bibinfo{author}{Z.~Qin}, \bibinfo{author}{K.~Ren}, \bibinfo{author}{W.~Ma},
\newblock \bibinfo{title}{A large-scale pretrained deep model for phishing url detection},
\newblock in: \bibinfo{booktitle}{ICASSP 2023-2023 IEEE International Conference on Acoustics, Speech and Signal Processing (ICASSP)}, \bibinfo{organization}{IEEE}, \bibinfo{year}{2023}, pp. \bibinfo{pages}{1--5}.
\bibitem[{Maneriker et~al.(2021)Maneriker, Stokes, Lazo, Carutasu, Tajaddodianfar, and Gururajan}]{maneriker2021urltran}
\bibinfo{author}{P.~Maneriker}, \bibinfo{author}{J.~W. Stokes}, \bibinfo{author}{E.~G. Lazo}, \bibinfo{author}{D.~Carutasu}, \bibinfo{author}{F.~Tajaddodianfar}, \bibinfo{author}{A.~Gururajan},
\newblock \bibinfo{title}{Urltran: Improving phishing url detection using transformers},
\newblock in: \bibinfo{booktitle}{MILCOM 2021-2021 IEEE Military Communications Conference (MILCOM)}, \bibinfo{organization}{IEEE}, \bibinfo{year}{2021}, pp. \bibinfo{pages}{197--204}.
\bibitem[{Lin(2022)}]{lin2022analysis}
\bibinfo{author}{R.~Lin},
\newblock \bibinfo{title}{Analysis on the selection of the appropriate batch size in cnn neural network},
\newblock in: \bibinfo{booktitle}{2022 International Conference on Machine Learning and Knowledge Engineering (MLKE)}, \bibinfo{organization}{IEEE}, \bibinfo{year}{2022}, pp. \bibinfo{pages}{106--109}.
\bibitem[{Loshchilov and Hutter(2017)}]{loshchilov2017decoupled}
\bibinfo{author}{I.~Loshchilov}, \bibinfo{author}{F.~Hutter},
\newblock \bibinfo{title}{Decoupled weight decay regularization},
\newblock \bibinfo{journal}{arXiv preprint arXiv:1711.05101}  (\bibinfo{year}{2017}).
\bibitem[{Zhuang et~al.(2022)Zhuang, Liu, Cutkosky, and Orabona}]{zhuang2022understanding}
\bibinfo{author}{Z.~Zhuang}, \bibinfo{author}{M.~Liu}, \bibinfo{author}{A.~Cutkosky}, \bibinfo{author}{F.~Orabona},
\newblock \bibinfo{title}{Understanding adamw through proximal methods and scale-freeness},
\newblock \bibinfo{journal}{arXiv preprint arXiv:2202.00089}  (\bibinfo{year}{2022}).
\bibitem[{Holzinger et~al.(2022)Holzinger, Saranti, Molnar, Biecek, and Samek}]{holzinger2022explainable}
\bibinfo{author}{A.~Holzinger}, \bibinfo{author}{A.~Saranti}, \bibinfo{author}{C.~Molnar}, \bibinfo{author}{P.~Biecek}, \bibinfo{author}{W.~Samek},
\newblock \bibinfo{title}{Explainable ai methods-a brief overview},
\newblock in: \bibinfo{booktitle}{International workshop on extending explainable AI beyond deep models and classifiers}, \bibinfo{organization}{Springer}, \bibinfo{year}{2022}, pp. \bibinfo{pages}{13--38}.
\bibitem[{Ribeiro et~al.(2016)Ribeiro, Singh, and Guestrin}]{ribeiro2016should}
\bibinfo{author}{M.~T. Ribeiro}, \bibinfo{author}{S.~Singh}, \bibinfo{author}{C.~Guestrin},
\newblock \bibinfo{title}{" why should i trust you?" explaining the predictions of any classifier},
\newblock in: \bibinfo{booktitle}{Proceedings of the 22nd ACM SIGKDD international conference on knowledge discovery and data mining}, \bibinfo{year}{2016}, pp. \bibinfo{pages}{1135--1144}.
\bibitem[{Chen et~al.(2024)Chen, Yin, Chen, and Wu}]{chen2024hybrid}
\bibinfo{author}{A.~Chen}, \bibinfo{author}{H.-L. Yin}, \bibinfo{author}{Z.-B. Chen}, \bibinfo{author}{S.~Wu},
\newblock \bibinfo{title}{Hybrid quantum-inspired resnet and densenet for pattern recognition with completeness analysis},
\newblock \bibinfo{journal}{arXiv preprint arXiv:2403.05754}  (\bibinfo{year}{2024}).
\bibitem[{Uddin et~al.(2022)Uddin, Shahriar, Haque, and Sarker}]{uddin2022cyber}
\bibinfo{author}{M.~A. Uddin}, \bibinfo{author}{K.~T. Shahriar}, \bibinfo{author}{M.~M. Haque}, \bibinfo{author}{I.~H. Sarker},
\newblock \bibinfo{title}{Cyber-attack detection through ensemble-based machine learning classifier},
\newblock in: \bibinfo{booktitle}{International Conference on Machine Intelligence and Emerging Technologies}, \bibinfo{organization}{Springer}, \bibinfo{year}{2022}, pp. \bibinfo{pages}{386--396}.
\bibitem[{Sarker(2024)}]{sarker2024llm}
\bibinfo{author}{I.~H. Sarker},
\newblock \bibinfo{title}{Llm potentiality and awareness: a position paper from the perspective of trustworthy and responsible ai modeling},
\newblock \bibinfo{journal}{Discover Artificial Intelligence} \bibinfo{volume}{4} (\bibinfo{year}{2024}) \bibinfo{pages}{40}.

\end{thebibliography}


\end{document}